\documentclass[11pt]{article}

\usepackage[preprint]{acl}

\usepackage{times}
\usepackage{latexsym}

\usepackage[T1]{fontenc}

\usepackage[utf8]{inputenc}

\usepackage{microtype}

\usepackage{inconsolata}

\usepackage{graphicx}
\usepackage{subcaption}

\usepackage{multirow}
\usepackage{makecell}
\usepackage[export]{adjustbox}
\usepackage{amsmath}
\usepackage{cleveref}
\usepackage{booktabs}

\newcommand{\llava}[0]{LLaVA-NeXT}
\newcommand{\llavaone}[0]{LLaVA-OneVision}
\newcommand{\molmo}[0]{Molmo}
\newcommand{\qwen}[0]{Qwen2-VL}
\newcommand{\clip}[0]{CLIP}
\newcommand{\llavas}[0]{LLaVA-NeXT }
\newcommand{\llavaones}[0]{LLaVA-OneVision }
\newcommand{\molmos}[0]{Molmo }
\newcommand{\qwens}[0]{Qwen2-VL }
\newcommand{\clips}[0]{CLIP }

\title{Synthetic Stimuli, Real Gains: \\Rethinking VLM Fine-Tuning Through Fully Controlled Data Generation}

 \author{Massimo Rizzoli, Simone Alghisi, Seyed Mahed Mousavi, Giuseppe Riccardi \\
         Signals and Interactive Systems Lab, University of Trento, Italy \\
         \texttt{\{massimo.rizzoli,giuseppe.riccardi\}@unitn.it }
         }

\begin{document}
\maketitle
\begin{abstract}
Performance gains of Vision Language Models (VLMs) obtained by fine-tuning are generally based on ad hoc data collection and annotation of real-world scenes. Despite the improvements, this process is often prone to biases, errors, and distribution imbalance, resulting in overfitting and imbalanced performance.
Although a few studies have explored synthetic data generation, they typically lack control over data distribution and annotation quality.
In this work, we re-evaluate the potential of model fine-tuning by exploring a fully controlled data generation and annotation pipeline, obtaining bias-free data with balanced distribution and clean annotations. Using the spatial reasoning task of identifying the absolute position of an object as a use case, we fine-tune state-of-the-art VLMs and conduct exhaustive evaluations on both synthetic and real-world benchmarks, including transferability to real-world scenes. Our experiments reveal two key findings: 1) fine-tuning on balanced data yields uniform performance across the visual scene and mitigates common biases with as few as 130 samples; and 2) fine-tuning on synthetic stimuli improves performance by 13\% on real-world data (COCO), outperforming models fine-tuned on the full COCO train set.\footnote{We release all materials: \url{https://github.com/sislab-unitn/synth-stimuli-real-gains}}
\end{abstract}

\section{Introduction}
\label{sec:intro}

Vision-Language Models (VLMs) have demonstrated competitive performance across a variety of downstream reasoning tasks, including visual question answering \cite{Goyal_2017_CVPR, Chen_2024_CVPR_SpatialVLM, Deitke_2025_CVPR_molmo_pixmo}, spatial reasoning \cite{krishna2017visual_genome, thrush2022winoground, yuksekgonul2023vlms_bagsofwords_ARO}, counting \cite{Acharya_Kafle_Kanan_2019_tallyqa, 10376915_clip_count_ten}, and visual scenes understanding \cite{fu-etal-2023-generate_then_select, cheng-etal-2024-from_least_to_most}.
To improve performance on these tasks, the prevailing approach is to collect task-specific annotated datasets from real-world scenarios, fine-tune the model on these data, and evaluate it on benchmarks built from similar distributions \cite{10.1007/978-3-031-73337-6_9_BLINK, Yue_2024_CVPR_MMMU}.
This pipeline has become the de facto paradigm for adapting and assessing VLMs in downstream tasks.
However, despite satisfactory benchmark performance, VLMs still exhibit severe limitations in understanding the structure and semantics of visual scenes \cite{kamath-etal-2023-whats_up_with_vlms, rudman-etal-2025-forgotten_polygons, rizzoli-etal-2025-civet}. Therefore, the improvement does not necessarily reflect enhanced generalization, as it may be driven by random or spurious correlations \cite{10378352_waffling, esfandiarpoor-etal-2024-clip_could_talk}.

A close inspection of the data used to improve (fine-tune) and evaluate (benchmark) VLMs' performance reveals annotation errors, distribution imbalance, and strong scene biases \cite{Acharya_Kafle_Kanan_2019_tallyqa, Kirillov_2023_ICCV, schuhmann2021laion}. As a result, both fine-tuning and evaluation reinforce each other’s limitations, giving the illusion of improvement while masking fundamental weaknesses in visual reasoning. Models fine-tuned on collected data often learn to associate task success with spurious cues, such as object co-occurrence or central positioning rather than generalization; meanwhile, as the benchmarks are constructed from the same biased distributions, evaluation rewards the models for reproducing dataset-specific shortcuts instead of robust understanding \cite{NEURIPS2024_2f8ee6a3_MMMU_issue, Rahmanzadehgervi_2024_ACCV_VLMs_blind}.

The current limitations in VLM understanding may result in catastrophic errors, especially in real-world deployment, where conditions differ from training.
For instance, a model might learn to detect pedestrians only when they appear near the image center, and fail when they occur elsewhere.
This highlights the need for a training and evaluation process that promotes task competence regardless of variability in irrelevant aspects, such as object color, shape, or position.

Recent studies have attempted to move beyond performance metrics, probing VLMs’ ability to reason about visual properties and relations \cite{Peng_2024_CVPR_SPEC, rudman-etal-2025-forgotten_polygons, chen2025why_spatial_reasoning_hard}.
These efforts highlight that benchmark results often conceal poor structural understanding and sensitivity to confounders.
However, these studies remain limited by partial coverage and remaining biases in their data, preventing a systematic analysis of how VLMs acquire and generalize spatial knowledge.
This highlights the need for systematic, controlled, and exhaustive datasets that enable the isolation of reasoning from spurious correlations.
In this work, we study the role of fully controlled data generation and annotation for fine-tuning VLMs, using the spatial reasoning task of identifying the absolute position of an object \cite{Peng_2024_CVPR_SPEC, rizzoli-etal-2025-civet} as a use case.
We frame our study around two central research questions.

\textbf{RQ1 (Assessment): Can controlled synthetic fine-tuning improve VLMs?}
Current training pipelines often expose models to dataset biases, annotation errors, and distribution imbalance.
We construct an exhaustive and balanced dataset to isolate performance from spurious cues and identify models' limitations.
For this purpose, we comprehensively synthesize object attributes such as color, shape, size, and position.
Focusing on the absolute position task as a use case, we fine-tune state-of-the-art VLMs and evaluate their ability to generalize across object configurations, measuring whether controlled training conditions enhance their spatial reasoning capabilities.

\textbf{RQ2 (Transfer): Do improvements learned from controlled synthetic data transfer to real-world scenes?}
While synthetic data enables controlled, exhaustive, and error-free coverage, models are required to perform reliably on real-world images.
To assess transferability, we evaluate VLMs fine-tuned on the synthetic dataset in an unmatched setting.
We construct a real-world dataset for the same downstream task, starting from COCO \cite{10.1007/978-3-319-10602-1_48_COCO}.
We further assess whether fine-tuning on synthetic data provides benefits over fine-tuning directly on real-world data by comparing transfer performance in the unmatched setting with a matched setting where models are fine-tuned and evaluated on real-world data.
This setup allows us to assess whether the acquired spatial reasoning skills of identifying the absolute position of an object extend beyond synthetic stimuli and enhance reliability in real-world scenarios.

Together, these research questions guide our investigation into how controlled synthetic data can enhance fine-tuning outcomes and transferability of VLMs.
Our experiments show that fine-tuning on controlled synthetic data improves model performance and transfers effectively to real-world settings.
Notably, improvements are most pronounced where models were previously biased.
Interestingly, fine-tuning on the entire COCO training set degrades performance, suggesting that more data is not always beneficial.
Moreover, while fine-tuning on a balanced subset of COCO training data (matched setting) also improves overall performance, it introduces biases such as failing to learn specific positions (e.g., center), and does not consistently achieve the robustness of our synthetic approach.


\section{Related Work}

\textbf{Scene Understanding} Improving the performance of VLMs via fine-tuning on task-specific data has been applied across diverse domains, including mathematical reasoning \cite{10.1007/978-3-031-73242-3_10_mathverse, shi-etal-2024-math-llava, gao2025gllava, zhang2025mavis}, visual relationship understanding \cite{NEURIPS2024_c2e06513_flexible_relation_segmentation}, scene graph construction \cite{Park_2025_CVPR_SVG}, spatial reasoning \cite{ogezi-shi-2025-spare, ning2025enhancing_spatial_reas_segm}, visual reasoning \cite{cheng-etal-2024-from_least_to_most}, and shape recognition \cite{rudman-etal-2025-forgotten_polygons}.
However, most studies inherit the issues of real-world data, while synthetic approaches often lack control over distribution and rely on annotations from generative models prone to hallucination.

\begin{figure*}[t]
    \centering
    \includegraphics[width=1.00\linewidth]{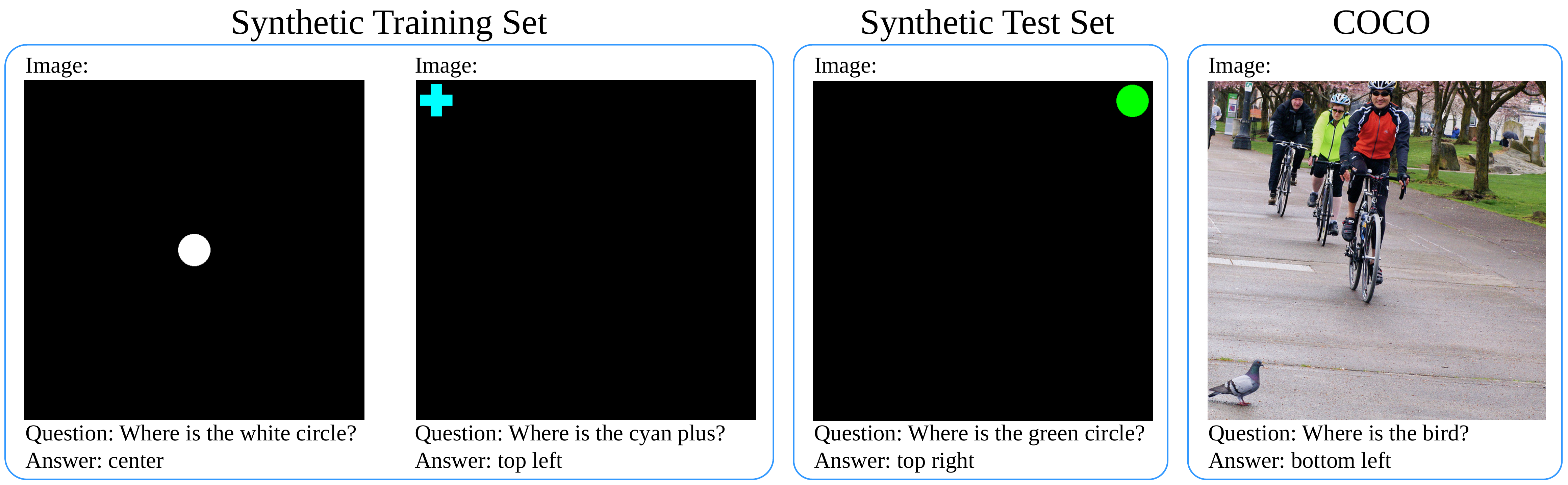}
    \caption{
        \textbf{Data samples.}
        Synthetic data consists of exhaustive sets of image-question pairs about a single object on a black background.
        Objects in the Synthetic Test Set are of color-shape combinations unseen in the Synthetic Training Set.
        The COCO Sets are obtained from the original COCO dataset by asking questions only for objects being the only instance of their category in a given image to avoid ambiguous questions.
    }
    \label{fig:data}
\end{figure*}

\textbf{Synthetic Data Generation} Recent studies have resorted to synthetic data to cope with issues related to real-world data. \citet{Johnson_2017_CVPR_CLEVR} aimed at avoiding annotation errors via deterministic scene generation.
Cut-and-paste approaches \cite{dwibedi2017cut_paste_learn, ghiasi2021simple_copypaste} augment images with objects extracted from others, but their reliance on image segmentation limits control over the content of the image.
SPEC \cite{Peng_2024_CVPR_SPEC} uses a diffusion-based approach to generate objects and background for the absolute position task.
Nevertheless, their approach suffers from hallucinations and inconsistencies \cite{NEURIPS2024_f29369d1_understanding_hallucination_diffusion, 10.1007/978-3-031-73004-7_6_structural_hallucination_diffusion}.
Similar issues are present in approaches that synthesize data for fine-tuning \cite{Li_2025_CVPR_sparcl_enhance_vlms_synthetic, Park_2025_CVPR_SVG} or for evaluation \cite{sbrolli2026autocomp} via generative models.
\citet{Wang_2025_CVPR_embodied_scene_underst_metavqa} generate synthetic scene and QA annotations, reducing labeling errors, but not addressing label imbalance. Other studies generate scenes consisting of geometric shapes \cite{Rahmanzadehgervi_2024_ACCV_VLMs_blind, rudman-etal-2025-forgotten_polygons, rizzoli-etal-2025-civet}, enabling systematic evaluation by isolating task-relevant factors and marginalizing irrelevant properties.
In a related study, \citet{kamath-etal-2023-whats_up_with_vlms} proposed a dataset of real-world images obtained by physically constructing scenes with controlled perturbations, which, while interesting, imposes limited scalability due to setup cost and time.


\section{Approach}
We investigate how a fully controlled data generation pipeline can improve fine-tuning outcomes via the \emph{Absolute Position} task, formulated as Visual Question Answering (VQA) over a $3\times3$ grid.
To disentangle task performance from dataset artifacts, we construct controlled synthetic datasets with exhaustive and balanced coverage using CIVET \cite{rizzoli-etal-2025-civet}. We then assess transferability by evaluating the performance of the fine-tuned model on COCO (unmatched), and compare against fine-tuning on real-world data from the same distribution (matched).

\subsection{Task: Absolute Position}
\label{sec:methods:task}
This task requires identifying in which of nine equally sized regions of an image a target object is located.
Each image is divided into a $3\times3$ grid representing nine possible locations: \emph{top left}, \emph{top center}, \emph{top right}, \emph{center left}, \emph{center}, \emph{center right}, \emph{bottom left}, \emph{bottom center}, and \emph{bottom right}. For each image, we generate a closed-ended VQA sample asking for the location of a specific object, e.g., \textit{``Where is the red square?''}. The nine grid locations are presented as answer options, and their order is randomized to prevent positional bias. This task setup follows recent work on spatial reasoning in VLMs \cite{Peng_2024_CVPR_SPEC, rizzoli-etal-2025-civet}.

\subsection{Dataset Construction}
\label{sec:dataset}
Using the CIVET framework \cite{rizzoli-etal-2025-civet}, we generate all synthetic data ensuring exhaustive coverage, balanced distributions, and the absence of annotation errors or sampling bias.
We use synthetic data to isolate performance from confounding factors, while real-world data enable us to test transferability in an unmatched setting. Therefore, we complement these synthetic datasets with a version of the same task built from the COCO dataset \cite{10.1007/978-3-319-10602-1_48_COCO}.
Data examples are illustrated in \Cref{fig:data}, and additional details on the data construction are reported in §\ref{app:dataset_construction}.

\textbf{Synthetic Test Set} We first build an exhaustive synthetic evaluation dataset to measure spatial reasoning.
Each image contains a single object on a uniform black background.
We systematically vary four object attributes (i.e., \emph{color}, \emph{shape}, \emph{size}, and \emph{position}), generate an image for each combination, and obtain a balanced dataset of 3,888\footnote{Computed as $6 \text{ colors} \times 4 \text{ shapes} \times 2 \text{ sizes} \times 81 \text{ fine-grained positions (i.e., } 9 \times 9 \text{ cells)}$} VQA samples following the formulation in \Cref{sec:methods:task}.

\begin{figure*}[h!]
    \centering
    \includegraphics[width=1.00\linewidth]{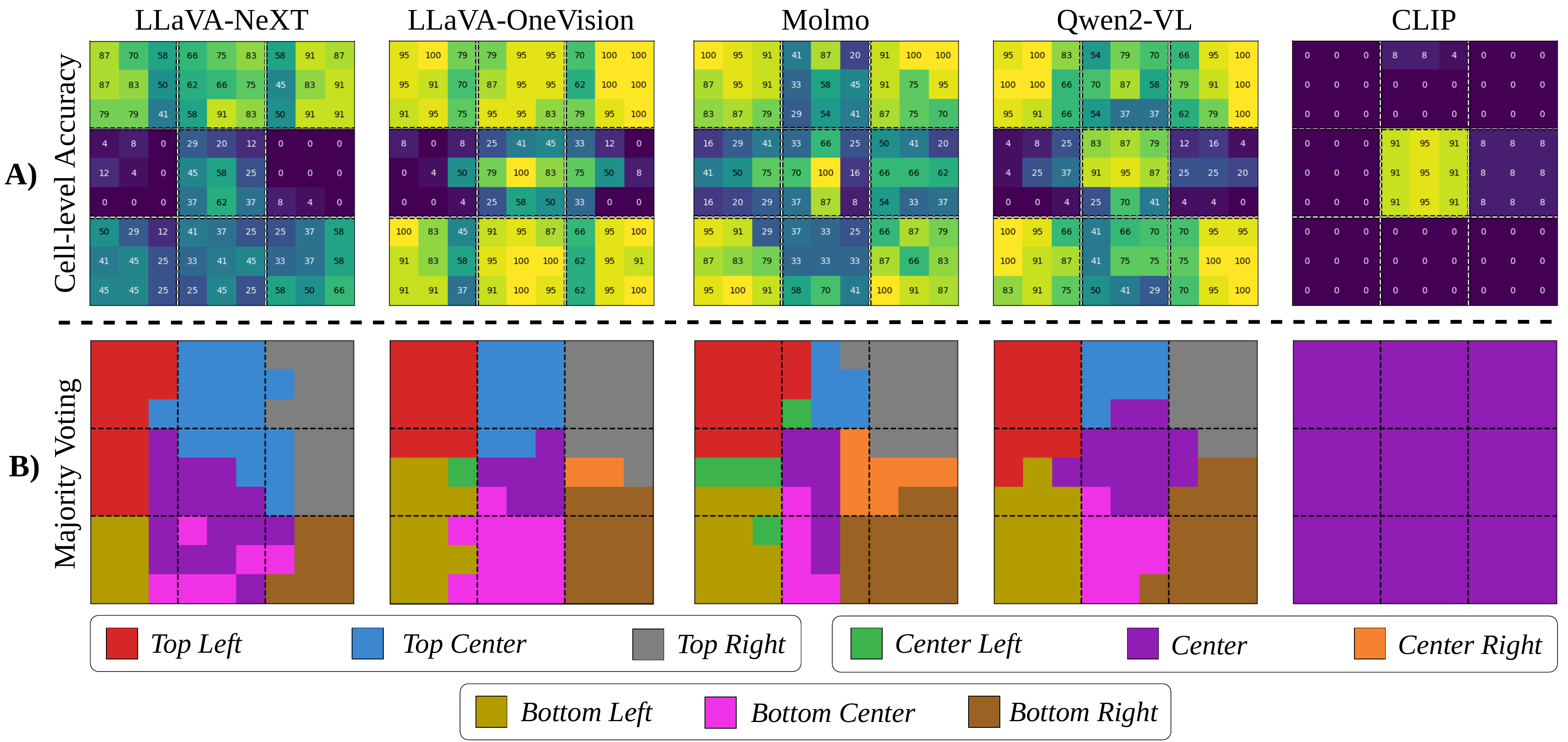}
    \caption{
        \textbf{A) Cell-Level Accuracy} and \textbf{B) Spatial Responses Map of VLM.} A) Accuracy on the \emph{Absolute Position} task averaged over object variations across a fine-grained grid of $9\times9$ cells shows pronounced positional biases before fine-tuning.
        The models perform best in upper regions, with consistently low accuracy in center-left and center-right cells; \clips exhibits an extreme central bias, failing elsewhere.
        B) Majority-vote answers reveal how models internally distort spatial structure.
    }
    \label{fig:cell_level_acc_and_repr_baseline}
\end{figure*}

\textbf{Synthetic Training Set}
To study whether controlled synthetic data can improve VLMs' spatial reasoning, we construct a training dataset with the same structure as the evaluation data but distinct color-shape combinations to avoid overlap (i.e., white \emph{<shape>} or \emph{<color>} plus).
The resulting dataset comprises 1,620\footnote{Computed as $(6 \text{ colored plusses} + 4 \text{ white shapes}) \times 2 \text{ sizes} \times 81 \text{ positions}$} image-question pairs, balanced across all positions.
We keep 80\% (1296) of the dataset for training and 20\% (324) for validation.
This configuration encourages the model to learn spatial reasoning independently of specific object shape or color cues, enabling an error-free analysis of controlled fine-tuning effects in both synthetic and unmatched real-world settings.

\textbf{Real-World Dataset}
To assess transferability to real-world data, we construct training and test datasets starting respectively from the train and validation\footnote{Test annotation is not provided.} splits of COCO.
For each image, we generate VQA samples that query the position of categories with a unique instance to avoid unambiguous questions (e.g., \textit{``Where is the person?''}).
We obtain a training set of 161,086 image-question pairs, a validation set of 40,272 pairs, and a test set of 8,548 pairs.
While remaining consistent with the synthetic setup, the dataset captures real-world variability (e.g., multiple objects, diverse layouts, and non-square aspect ratios) and enables systematic analysis of how absolute position is learned in synthetic settings and transfers to real-world scenes.

\subsection{Vision-Language Models}
We evaluate five VLMs representative of the main dual-encoder and encoder-decoder architectures, allowing us to study balanced synthetic fine-tuning across design families.
\clips \cite{radford2021clip} is a dual-encoder model that learns aligned image and text representations through contrastive training. We include \clips as a baseline and because its vision encoder serves as the foundation for several subsequent encoder-decoder VLMs.
\llavas 7B \cite{liu2024llavanext} builds on \clips by projecting visual features into the embedding space of a Large Language Model (LLM) through a learned projection layer.
\molmos 7B \cite{Deitke_2025_CVPR_molmo_pixmo} follows a similar design but fine-tunes the entire architecture.
The more recent \qwens 7B \cite{wang2024qwen2} and \llavaones 8B \cite{an2025llava_onevision} directly process images of varying resolutions without cropping or resizing.


\section{Experiments}
\label{sec:experiments}

\begin{figure*}[t]
\centering
\begin{subfigure}[t]{0.46\linewidth}
    \centering
    \includegraphics[width=\linewidth]{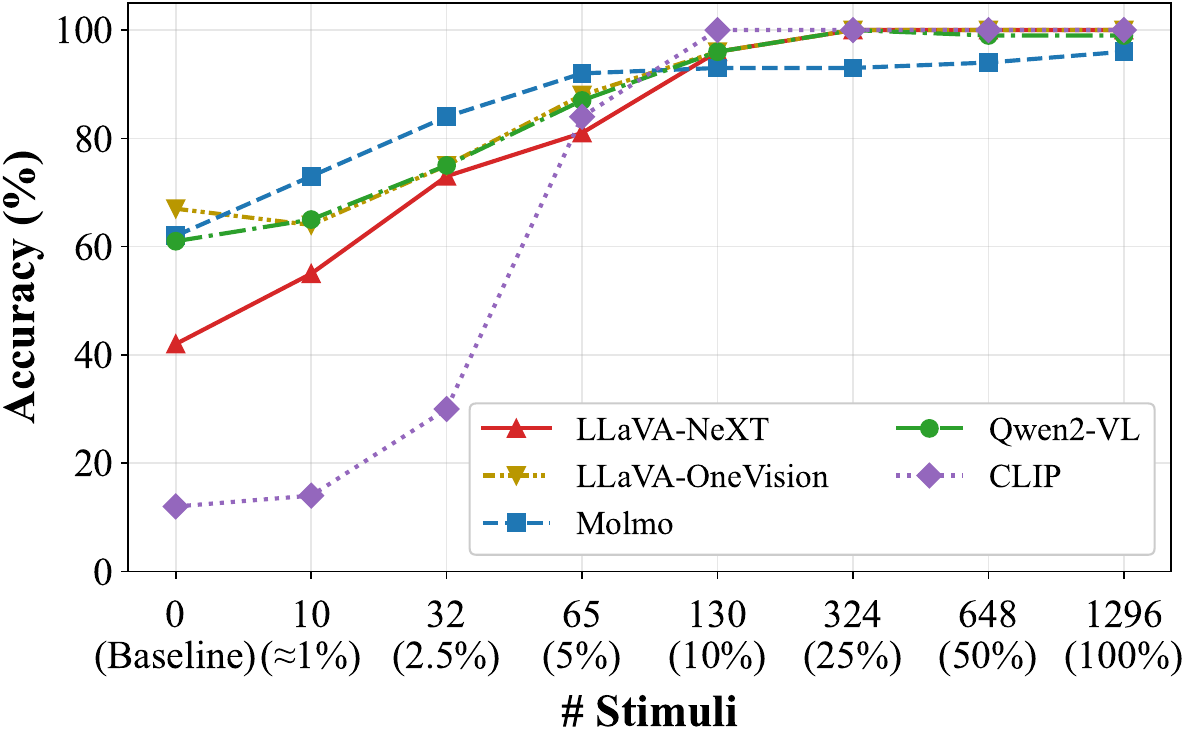}
    \caption{Synthetic Test Set}
    \label{fig:scaling_data_synth}
\end{subfigure}
\hspace{0.04\linewidth}
\begin{subfigure}[t]{0.46\linewidth}
    \centering
    \includegraphics[width=\linewidth]{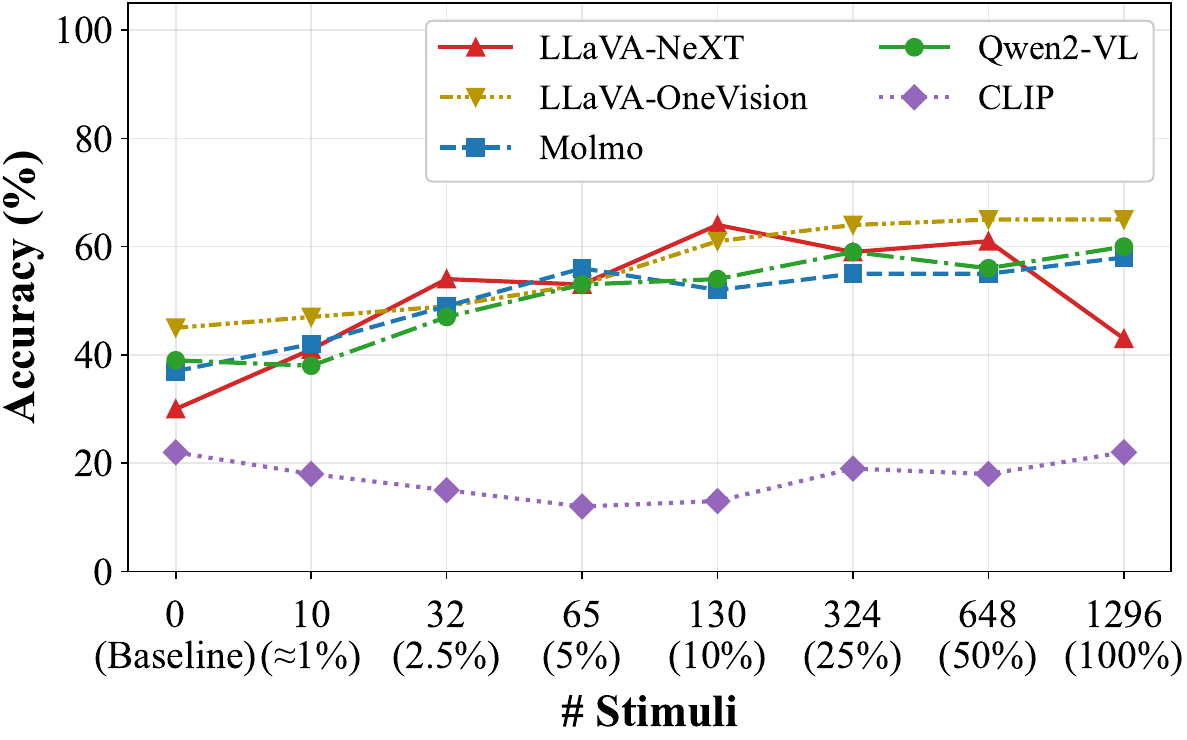}
    \caption{COCO Test Set}
    \label{fig:scaling_data_COCO_test}
\end{subfigure}
\caption{
    \textbf{Effect of synthetic dataset scale.} Accuracy on the \emph{Absolute Position} task as a function of the number of synthetic training stimuli.
    \textbf{(a)} Effect on the Synthetic Test Set.
    \textbf{(b)} Effect on the COCO Test Set.
}
\end{figure*}
\subsection*{RQ1: Can Controlled Synthetic Fine-Tuning Improve VLMs?}
To investigate the effect of controlled synthetic fine-tuning, we begin by evaluating the spatial reasoning behavior of base models in the absolute position task.
We then evaluate how fine-tuning on balanced synthetic data reshapes and improves these behaviors.
\\\textbf{A. Cell-Level Accuracy} To evaluate the spatial biases of base models before fine-tuning, we analyze both their fine-grained positional accuracy and spatial structure (\Cref{fig:cell_level_acc_and_repr_baseline}). For cell-level accuracy (\Cref{fig:cell_level_acc_and_repr_baseline}A), we subdivide the $3\times3$ region grid into a finer $9\times9$ cell layout and compute the mean accuracy over all object variations within each cell.

\textbf{Results:} The results (\Cref{fig:cell_level_acc_and_repr_baseline}) indicate that all models exhibit strong spatial biases prior to fine-tuning.
\llavaones, \molmo, and \qwens perform best in the upper and lower regions, while \llavas achieves high accuracy only in the upper regions, and \clips achieves high accuracy only in the central region and fails elsewhere.
A consistent weakness emerges in the center-left and center-right regions, and near region boundaries, where all VLMs struggle.
While \llavaone, \molmo, and \qwens show better coverage, no model achieves uniform spatial performance.
\\\textbf{B. Spatial Responses Map of VLMs}
We further analyze the models’ answers regarding the absolute position through majority-vote (\Cref{fig:cell_level_acc_and_repr_baseline}B).
For each cell, we aggregate answers across all object variations, and color-code the cells according to the most frequent answer position.
This visualization exposes how models' answers ``remap'' the spatial grid before fine-tuning.

\textbf{Results:}
\Cref{fig:cell_level_acc_and_repr_baseline}B shows that llavas over-represents the upper half, with many central cells misclassified as upper positions, and the bottom center collapsed into the center.
While more symmetric, \llavaones and \qwens exhibit strong vertical compression, with top and bottom regions substituting the central-left and central-right regions.
\molmos produces a more coherent layout but compresses the central regions both horizontally and vertically.
\clips degenerates into answering only with center, confirming its extreme central bias observed in the cell-level accuracy.
Together, these two analyses reveal that VLMs encode strong spatial bias towards top regions, and fail to represent intermediate regions.
This highlights the necessity of fine-tuning on controlled synthetic data to eliminate such biases and foster accurate spatial representations.
\\\textbf{C. Fine-Tuning} We investigate whether fine-tuning on balanced synthetic data can enhance the spatial reasoning capabilities of VLMs in the absolute position task.
Fine-tuning details are reported in §\ref{app:fine_tuning_details}.
We fine-tune the models on the balanced synthetic dataset and compute the mean accuracy across five runs (detailed results and standard deviation are reported in §\ref{app:fine_tuning_results}).

\textbf{Results:}
While the base models achieve at best 67\% accuracy, fine-tuning consistently improves spatial reasoning across all models, achieving near-perfect accuracy ($\geq$ 96\%) and minimal variance across runs.
Overall, these findings validate our first research question (RQ1), i.e. fine-tuning on balanced synthetic data substantially enhances performance on the absolute position task while maintaining robustness across training runs.
\\\textbf{D. Scaling Synthetic Data} We evaluate how progressively increasing the size of the synthetic training set affects model performance.

\textbf{Results:}
Across all models, accuracy increases rapidly with a small number of training stimuli (\Cref{fig:scaling_data_synth}). Most models reach near-optimal accuracy with only 10\% of the full set, after which performance plateaus, suggesting diminishing returns from additional data.
\molmos exhibits the fastest gains, achieving strong performance even with limited data, while \llavas, \llavaone, and \qwens improve more gradually but ultimately converge at a similar level.
\clips shows a different pattern, with minimal improvement at small scales followed by a sharp increase once sufficient samples are available, reflecting a greater dependence on data volume.
Overall, the results demonstrate that fine-tuning on a small, balanced subset of synthetic data is sufficient to achieve robust performance, highlighting the sample efficiency of fine-tuning on controlled synthetic data.

\subsection*{RQ2: Do Improvements Learned from Controlled Synthetic Data Transfer to Real-World Scenes?}
After observing improved performance by fine-tuning on controlled synthetic data, we investigate whether these improvements transfer to real-world data by evaluating models on COCO Absolute Position.
To probe transferability, we consider two complementary evaluation conditions: i) \textit{unmatched setting}, where models are fine-tuned on synthetic data and evaluated on COCO; and ii) \textit{a matched setting}, where models are fine-tuned and evaluated on COCO data. This comparison enables us to disentangle whether the benefits of exhaustive, bias-free synthetic training extend to uncontrolled real-world distributions.
\begin{table}[t]
    \centering
    \begin{adjustbox}{width=\columnwidth}
        \small
        \setlength{\tabcolsep}{5pt}
        \begin{tabular}{@{}llll@{}}
            \toprule
            \multirow{3}{*}{\textbf{Model}} & \multirow{3}{*}{{\makecell[c]{\textbf{Training Set}\\(\#Samples)}}} & \multicolumn{2}{c}{\multirow{2}{*}{\makecell[c]{\textbf{Test Set}\\ \textbf{Accuracy (\%)}}}} \\ \\
            \cmidrule{3-4}
             & & \textbf{Synthetic} & \textbf{COCO} \\
            \midrule

            \multirow{2}{*}{\llava}
              & Synthetic (1.3k)                        & 100 \textcolor{green!60!black}{($\uparrow$~58)} & 43 \textcolor{green!60!black}{($\uparrow$~13)} \\
              &  COCO \textit{Complete} (161k)                             & 0 \textcolor{red!60!black}{($\downarrow$~42)} & 0 \textcolor{red!60!black}{($\downarrow$~30)} \\

            \midrule
            \multirow{2}{*}{\llavaone}
              & Synthetic (1.3k)                        & 100 \textcolor{green!60!black}{($\uparrow$~33)} & 65 \textcolor{green!60!black}{($\uparrow$~20)} \\
              &  COCO \textit{Complete} (161k)                             & 11 \textcolor{red!60!black}{($\downarrow$~56)} & 26 \textcolor{red!60!black}{($\downarrow$~19)} \\

            \midrule
            \multirow{2}{*}{\molmo}
              & Synthetic (1.3k)                        & 96 \textcolor{green!60!black}{($\uparrow$~34)} & 58 \textcolor{green!60!black}{($\uparrow$~21)} \\
              &  COCO \textit{Complete} (161k)    & 4 \textcolor{red!60!black}{($\downarrow$~58)} & 6 \textcolor{red!60!black}{($\downarrow$~31)} \\

            \midrule
            \multirow{2}{*}{\qwen}
              & Synthetic (1.3k)                        & 99 \textcolor{green!60!black}{($\uparrow$~38)} & 60 \textcolor{green!60!black}{($\uparrow$~21)} \\
              &  COCO \textit{Complete} (161k)    & 9 \textcolor{red!60!black}{($\downarrow$~52)} & 20 \textcolor{red!60!black}{($\downarrow$~19)} \\

            \midrule
            \multirow{2}{*}{\clip}
              & Synthetic (1.3k)                        & 100 \textcolor{green!60!black}{($\uparrow$~88)} & 22 \textcolor{gray!70!black}{($=$)} \\
              &  COCO \textit{Complete} (161k)    & 11 \textcolor{red!60!black}{($\downarrow$~1)} & 36 \textcolor{green!60!black}{($\uparrow$~14)} \\

            \bottomrule
        \end{tabular}
    \end{adjustbox}
    \caption{
        \textbf{Cross-domain transfer to real-world data}.
        Accuracy (\%) on the \emph{Absolute Position} task for models fine-tuned on balanced Synthetic (1.3k), and COCO Complete (161k) test sets.
        Results are averaged over 5 runs.
        Arrows $($\textcolor{green!60!black}{$\uparrow$}$/$\textcolor{red!60!black}{$\downarrow$}$)$ indicate absolute increase and decrease in accuracy with respect to the \textit{Base Model}, while \textcolor{gray!70!black}{($=$)} denotes no change.
    }
    \label{tab:synth_vs_coco_train}
    \label{tab:synth_vs_coco_train_a}
\end{table}
\begin{figure*}[t]
    \centering
    \includegraphics[width=0.90\linewidth]{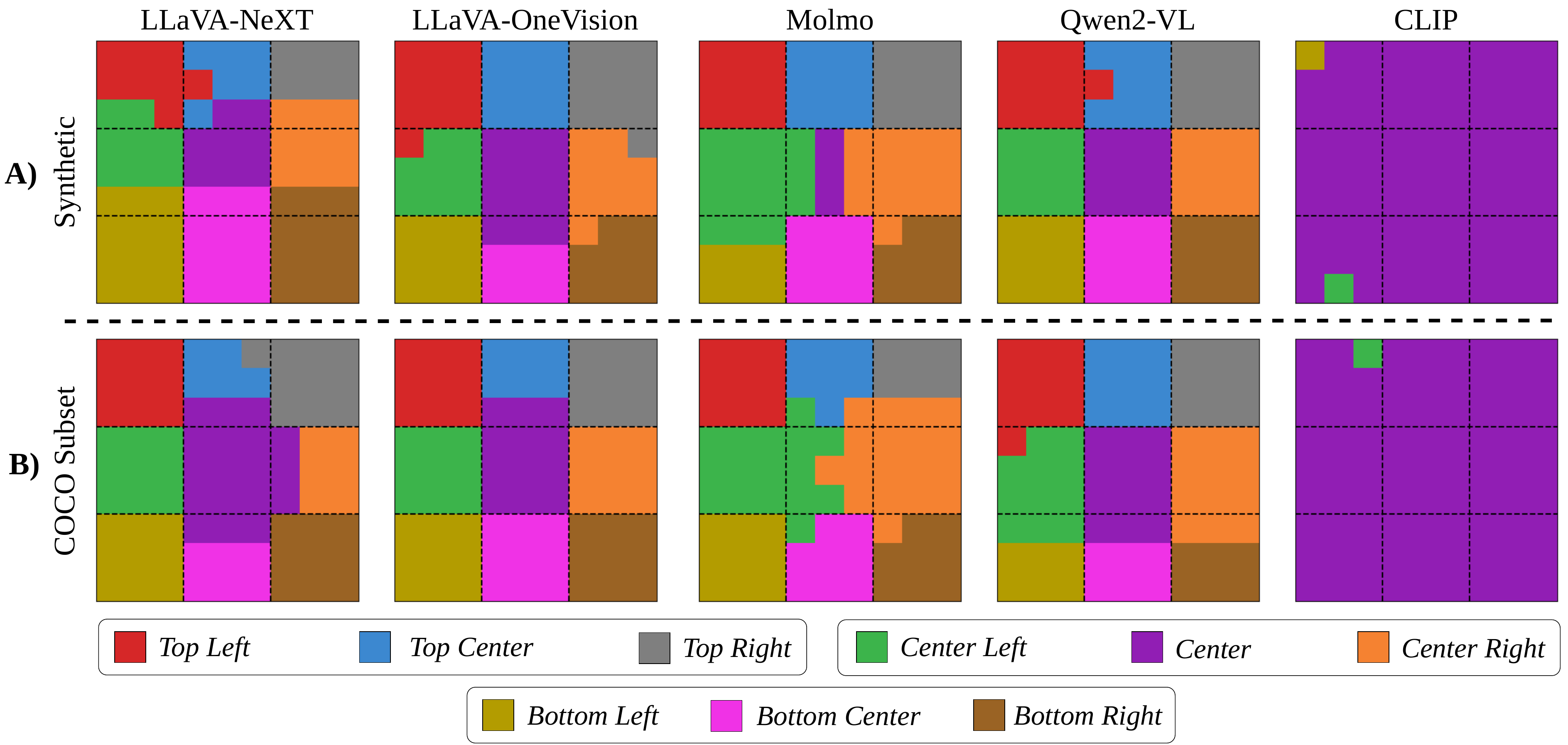}
    \caption{\textbf{Model answers on the COCO Absolute Position test set} after fine-tuning on different data sources (by majority voting); A) Models fine-tuned on synthetic data; and, B) Models fine-tuned on COCO Subset.}
    \label{fig:ft_synth_vs_ft_coco_repr}
\end{figure*}
\\\textbf{A. Cross-Domain Transfer} We evaluate the VLMs on the COCO Absolute Position test set to measure how effectively spatial reasoning learned from synthetic data transfers to real-world images.
Each model is fine-tuned on the synthetic training dataset and subsequently tested both on the synthetic and COCO benchmarks.
To compare the performance with the matched setting, we additionally fine-tune models on the complete COCO training set ($\sim$161k samples).

\textbf{Results} \Cref{tab:synth_vs_coco_train} summarizes model accuracy across matched (synthetic) and unmatched (COCO) evaluation settings.
Fine-tuning on the balanced synthetic dataset markedly improves performance on the absolute position task across all encoder-decoder models, not only on the matched synthetic test but also when transferring to real-world data.
\llavaone, \molmo, and \qwens each show gains of +20\% points or more on COCO, achieving around 60\% accuracy after synthetic fine-tuning. This indicates that models trained on controlled stimuli acquire transferable reasoning rather than overfitting to synthetic patterns.
Nevertheless, \clips fails to benefit from fine-tuning on synthetic data, suggesting a limitation of dual-encoder models.
In contrast, models fine-tuned on the full COCO training set (\Cref{tab:synth_vs_coco_train}) exhibit strong degradation (additional analysis is presented in §\ref{app:synth_to_coco_transfer}), with some models' performance dropping to near-zero accuracy, despite keeping the training procedure and hyperparameters the same as the synthetic fine-tuning.
\llavas generates empty strings as output, while \molmos often generates tokens from the set \emph{$\{$center, right, left$\}$}, resulting in invalid answers.
Instead, \llavaones and \qwens generate answers that are mostly valid, but often incorrect, reaching 50\% accuracy only in the center region, which is the most frequent in the data.
This suggests that naive fine-tuning on large-scale real-world data can inject noise and bias that hinder the learning of consistent spatial structure.
\begin{table}[t]
        \centering
        \small
        \setlength{\tabcolsep}{4pt}
        \begin{tabular}{@{}lp{1.85cm}l@{}}
            \toprule
            \multirow{2}{*}{\textbf{Model}} & \multicolumn{2}{c}{\textbf{Test Set} \textbf{Accuracy (\%)}} \\
            \cmidrule{2-3}
             & \textbf{Synthetic} & \textbf{COCO} \\
            \midrule

            \multirow{1}{*}{\llava}
              & 71 \textcolor{green!60!black}{($\uparrow$~29)} & 67 \textcolor{green!60!black}{($\uparrow$~37)} \\

            \multirow{1}{*}{\llavaone}
              & 77 \textcolor{green!60!black}{($\uparrow$~10)} & 64 \textcolor{green!60!black}{($\uparrow$~19)} \\

            \multirow{1}{*}{\molmo}
              & 80 \textcolor{green!60!black}{($\uparrow$~18)} & 45 \textcolor{green!60!black}{($\uparrow$~8)} \\

            \multirow{1}{*}{\qwen}
              & 80 \textcolor{green!60!black}{($\uparrow$~19)} & 61 \textcolor{green!60!black}{($\uparrow$~22)} \\

            \multirow{1}{*}{\clip}
              & 13 \textcolor{green!60!black}{($\uparrow$~1)} & 36 \textcolor{green!60!black}{($\uparrow$~14)} \\

            \bottomrule
        \end{tabular}
    \caption{
        \textbf{Fine-Tuning on balanced real-world data}.
        Accuracy (\%) on the \emph{Absolute Position} task for models fine-tuned on a COCO Subset (1.3k), balanced in terms of object category and position.
        Results are averaged over 5 runs.
        Arrows $($\textcolor{green!60!black}{$\uparrow$}$)$ indicate absolute increase in accuracy with respect to the \textit{Base Model}.
    }
    \label{tab:synth_vs_coco_train_b}
\end{table}
\\\textbf{B. Balanced Real-World Fine-Tuning}
To test whether data scale and imbalance hinder learning rather than the real-world setting, we construct a subset of COCO equivalent in size to our synthetic training set (i.e., 1,296 samples), balanced in object categories and positional distribution.

\textbf{Results:}
Interestingly, fine-tuning models on the balanced COCO Subset improves results and outperforms fine-tuning on the full COCO training set (\Cref{tab:synth_vs_coco_train_b}).
Notably, fine-tuning models on our controlled Synthetic training set (unmatched) or the balanced COCO Subset (matched) yields similar results when tested on COCO, but with the COCO Subset leading to more unstable performance, with much higher gains for \llavas and considerably lower for \molmo.
Overall, these results demonstrate that quality, balance, and control in training data outweigh sheer quantity.
\\\textbf{C. Data Scale and Transfer Efficiency} To understand how data quantity influences the performance to real-world settings, we progressively increase the number of synthetic training samples and evaluate model accuracy on the COCO test set.

\textbf{Results:}
Across VLM models, performance improves sharply even with a small fraction of the synthetic dataset, demonstrating the sample efficiency of balanced synthetic fine-tuning (\Cref{fig:scaling_data_COCO_test}).
With 10\% of the full synthetic data (130 samples), \llavas achieves its maximum transfer accuracy, and \llavaone, \molmo, and \qwens obtain most of their transfer improvement, after which performance plateaus.
In contrast to encoder-decoder VLMs, \clips remains largely insensitive to training size, with accuracy fluctuating around 20\%, suggesting that dual-encoder architectures do not effectively transfer from fine-tuning on synthetic data. Overall, these results highlight that balanced synthetic data achieves strong transfer with few samples, and that careful control and balance are far more beneficial than scale alone.
\\\textbf{D. Cell-Level Accuracy} To better understand how fine-tuning affects positional reasoning, we analyze cell-level accuracy and model answer patterns before and after fine-tuning.

\begin{figure}[t]
    \centering
    \includegraphics[width=0.90\linewidth]{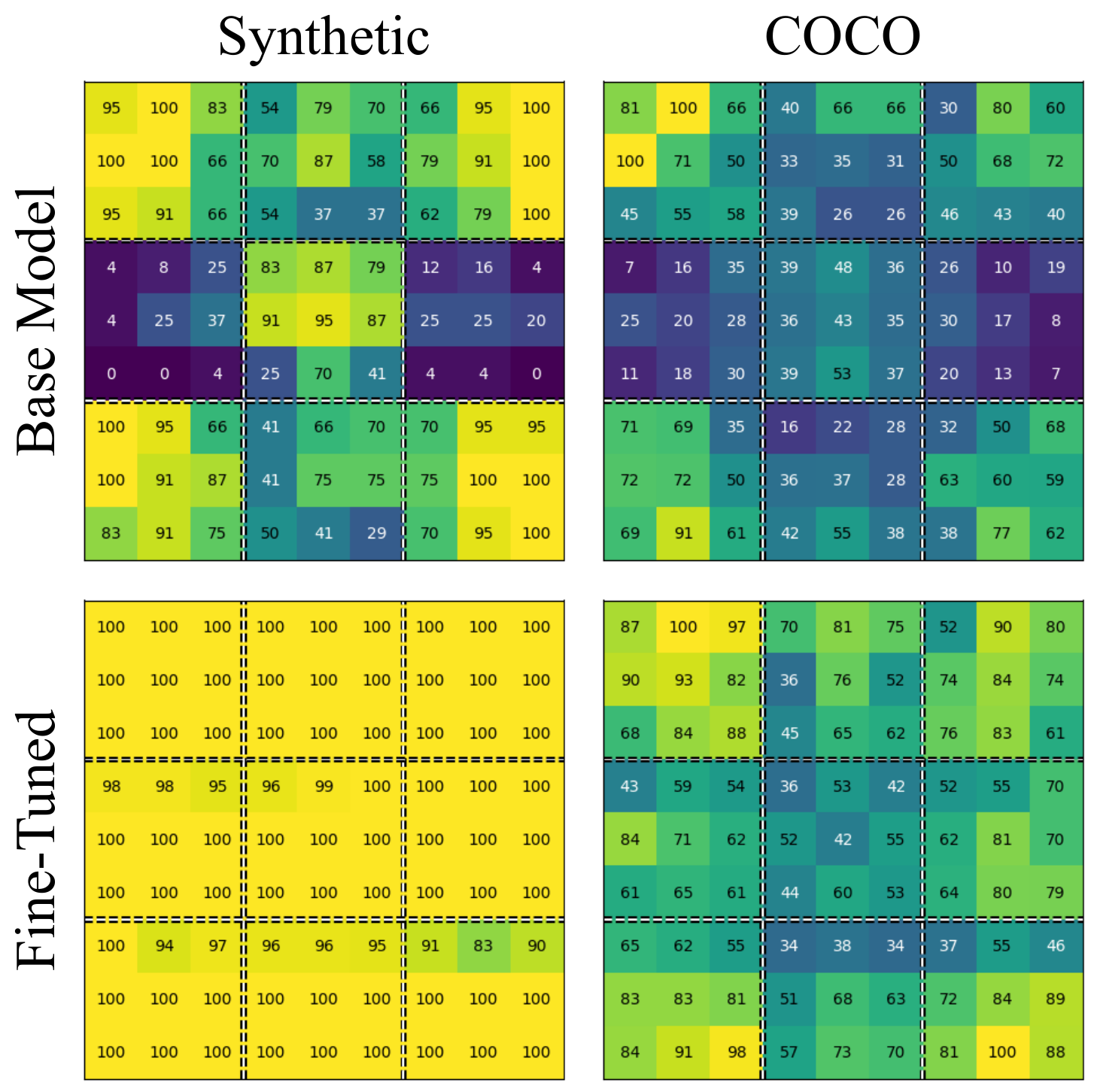}
    \caption{\textbf{Cell-level accuracy of \qwens 7B before and after fine-tuning} on synthetic data, on both the synthetic and COCO Absolute Position test sets.}
    \label{fig:improvements_synth_coco_qwen}
\end{figure}

\textbf{Results:}
Overall, these analyses demonstrate that fine-tuning on controlled synthetic data not only enhances positional accuracy but also refines spatial answers into coherent layouts. \Cref{fig:improvements_synth_coco_qwen} illustrates cell-level accuracy of \qwen, evaluated on both the synthetic and COCO test sets (other models are presented in §\ref{app:cell_level_accuracy}). Before fine-tuning, the model exhibits strong spatial biases, performing best in the upper and lower regions while struggling in the center-left and center-right. Fine-tuning on synthetic data improves performance as the accuracy becomes nearly uniform across all $9\times9$ cells, with the largest gains mostly where the base model performed worst.
Crucially, these improvements transfer to COCO, indicating that the model improved spatial reasoning capability for the absolute position rather than memorizing synthetic patterns.
\Cref{fig:ft_synth_vs_ft_coco_repr} further visualizes position answers on COCO after fine-tuning on different data sources.
Models fine-tuned on synthetic data (\Cref{fig:ft_synth_vs_ft_coco_repr}A) generally produce consistent and well-structured spatial partitions, whereas fine-tuning on the balanced COCO subset (\Cref{fig:ft_synth_vs_ft_coco_repr}B) can yield noisier and less regular layouts. Notably, in \molmos the \textit{center} region is effectively overwritten after COCO fine-tuning, indicating that real-world data may be more challenging to learn from, even when balanced.

\section{Ablation and Representation Analyses}
\label{sec:ablation_repr_analysis}
We further investigate the factors that influence the robustness and interpretability of VLMs after controlled fine-tuning.
\\\textbf{A. Scene Complexity \& Distractors}
As real-world scenes in COCO contain, on average, seven objects, we augment our synthetic dataset with distractor objects (details in §\ref{app:synth_with_distractors_construction}).
This allows us to systematically evaluate how increasing training scene complexity affects spatial reasoning over absolute position and transfer to real-world data.
We fine-tune each VLM on synthetic datasets containing one, three, or five distractors and evaluate them on the Synthetic and COCO Absolute Position test sets (additional results are reported in §\ref{app:distractors_results}).

\textbf{Results:}
\Cref{tab:distractors} shows that moderate visual clutter improves transfer to COCO for encoder-decoder VLMs. \llavas and \molmos benefit the most from adding three distractors, gaining +12\% and +3\% points, respectively on COCO. However, excessive clutter (five distractors) leads to diminishing or negative returns, suggesting that overly complex synthetic scenes can hinder learning and transfer. \qwens exhibits stable performance up to three distractors but slight degradation beyond that, indicating a similar saturation effect.
\llavaones shows improvement with distractors, but with reduced gains with respect to the clean set.
\clips remains largely unaffected, consistent with the limited transferability we observed.
Overall, these findings indicate that introducing moderate scene complexity during fine-tuning enhances robustness and transfer to real-world data.
\\\textbf{B. Layer-wise Representation Analysis}
To better understand how fine-tuning reshapes the internal representations of VLMs, we perform a layer-wise performance analysis \cite{fu2025hidden_plain_sight, alghisi2025de_re_constructing} before and after fine-tuning on our synthetic training set. For each layer of the LLM component, we extract the hidden representation corresponding to the final question token and train a linear SVM probe (3-fold cross-validation) to predict the absolute position label. This analysis allows us to localize where spatial reasoning emerges in the model hierarchy and how fine-tuning alters the encoding of spatial information (plots for each model are reported in §\ref{app:layerwise_results}).

\begin{table}[t]
  \centering
  \small
  \begin{adjustbox}{width=\columnwidth}

  \setlength{\tabcolsep}{5pt}
  \begin{tabular}{@{}llll@{}}
    \toprule
    \multirow{2}{*}{\textbf{Model}} &  \multirow{2}{*}{\textbf{Training Set}} & \multicolumn{2}{c}{\textbf{Test Set Accuracy (\%)}} \\\cmidrule{3-4}

                           & & \textbf{Synthetic} & \textbf{COCO} \\

    \midrule
   \multirow{3}{*}{\llava} &  {Synthetsubsetic (1.3k)}           & 100 \textcolor{green!60!black}{($\uparrow$~58)} & 42 \textcolor{green!60!black}{($\uparrow$~12)} \\
    &  \quad\textit{+3 Distractors} & 100 \textcolor{green!60!black}{($\uparrow$~58)}& 54 \textcolor{green!60!black}{($\uparrow$~24)}\\
    &  \quad\textit{+5 Distractors} & 81 \textcolor{green!60!black}{($\uparrow$~39)} & 48 \textcolor{green!60!black}{($\uparrow$~18)}\\
    \midrule
   \multirow{3}{*}{\llavaone} &  {Synthetic (1.3k)}           & 100 \textcolor{green!60!black}{($\uparrow$~33)} & 65 \textcolor{green!60!black}{($\uparrow$~20)} \\
    &  \quad\textit{+3 Distractors} & 100 \textcolor{green!60!black}{($\uparrow$~33)}& 60 \textcolor{green!60!black}{($\uparrow$~15)}\\
    &  \quad\textit{+5 Distractors} & 100 \textcolor{green!60!black}{($\uparrow$~33)} & 60 \textcolor{green!60!black}{($\uparrow$~15)}\\
    \midrule
    \multirow{3}{*}{\molmo} & {Synthetic (1.3k)}           & 96 \textcolor{green!60!black}{($\uparrow$~34)} & 57 \textcolor{green!60!black}{($\uparrow$~18)}\\
    & \quad\textit{+3 Distractors} & 95 \textcolor{green!60!black}{($\uparrow$~33)} & 60 \textcolor{green!60!black}{($\uparrow$~21)}\\
    & \quad\textit{+5 Distractors} & 97 \textcolor{green!60!black}{($\uparrow$~35)} & 65 \textcolor{green!60!black}{($\uparrow$~26)} \\
    \midrule
    \multirow{3}{*}{\qwen} & {Synthetic (1.3k)}           & 99 \textcolor{green!60!black}{($\uparrow$~38)}  & 58 \textcolor{green!60!black}{($\uparrow$~20)} \\
    & \quad\textit{+3 Distractors} & 93 \textcolor{green!60!black}{($\uparrow$~32)}  & 58 \textcolor{green!60!black}{($\uparrow$~20)} \\
    & \quad\textit{+5 Distractors} & 90 \textcolor{green!60!black}{($\uparrow$~29)}  & 54 \textcolor{green!60!black}{($\uparrow$~16)} \\
    \midrule
    \multirow{3}{*}{\clip} & {Synthetic (1.3k)}           & 100 \textcolor{green!60!black}{($\uparrow$~88)} & 22 \textcolor{gray!70!black}{($=$)} \\
    & \quad\textit{+3 Distractors} & 11 \textcolor{red!60!black}{($\downarrow$~1)} & 22 \textcolor{gray!70!black}{($=$)}\\
    & \quad\textit{+5 Distractors} & 11 \textcolor{red!60!black}{($\downarrow$~1)} & 28 \textcolor{green!60!black}{($\uparrow$~6)} \\
    \bottomrule
  \end{tabular}
  \end{adjustbox}
  \caption{
      \textbf{Effect of fine-tuning with distractors} on the \emph{Absolute Position} task.
      Results show the average accuracy (\%) across five runs.
      Arrows $($\textcolor{green!60!black}{$\uparrow$}$/$\textcolor{red!60!black}{$\downarrow$}$)$ indicate absolute increase and decrease in accuracy with respect to the \textit{Base Model}, while \textcolor{gray!70!black}{($=$)} denotes no change.
  }
  \label{tab:distractors}
\end{table}

\textbf{Results:}
On synthetic data, accuracy rapidly increases for all models in early layers (5$\sim$10) and saturates in the upper-middle layers.
In contrast, on COCO the same trend appears with a slower rise (saturation after layer 15), reflecting the increased visual and linguistic complexity of real-world scenes.
Together, these results indicate that fine-tuning on controlled synthetic data strengthens the internal representation of VLMs and that the learned representation for absolute position largely transfers to real-world settings, albeit with reduced confidence and stability.


\section{Conclusion}
We studied a controlled approach to fine-tune Vision-Language Models, showing that controlled data can improve the fine-tuning outcomes and transfer to real-world scenes on the spatial reasoning task of absolute position.
By systematically varying visual attributes and scene complexity, we isolated how models acquire and generalize spatial knowledge, revealing that the quality and balance of data matter more than scale.
Our analyses further demonstrated that controlled fine-tuning reshapes model representations in interpretable ways and promotes robustness across models and complex scenes.

Beyond the specific task of spatial reasoning, our findings suggest that synthetic data, when exhaustively designed and bias-free, can serve as a reliable tool for diagnosing, training, and benchmarking encoder-decoder models. Future work should investigate how controlled stimuli can be extended to other reasoning dimensions, such as relational, causal, and temporal understanding, and how such targeted fine-tuning might complement large-scale pretraining. Bridging synthetic precision with real-world richness offers a path towards VLMs that not only perform well but also reason reliably and transparently across visual domains.

\section*{Limitations}
Our controlled setup and exhaustive, bias-free analyses show that fine-tuning with controlled data is beneficial when tested on the spatial reasoning task of absolute position.
However, absolute position represents one aspect of spatial reasoning, and future work should investigate whether similar benefits extend to other spatial reasoning tasks, as well as to other reasoning dimensions such as relational, causal, and temporal understanding.
Moreover, due to computational and resource constraints, our experiments are limited to open-weight VLMs up to 8B parameters. Closed API-based models were also not considered, as they cannot be fine-tuned. Finally, the observed improvements are not present for CLIP, suggesting that the effectiveness of controlled fine-tuning may also depend on pre-training strategies and architectural design choices.

\bibliography{custom}

\clearpage
\appendix

\section{Dataset Construction}
\label{app:dataset_construction}
We report details on the construction of the datasets complementing \Cref{sec:dataset}.

\textbf{Synthetic Test Set}
We build an exhaustive synthetic evaluation dataset for the absolute position task.
Each image contains a single object on a uniform black background.
We systematically vary four object attributes: color, shape, size, and position.
We use six colors (red, green, blue, cyan, magenta, yellow), four shapes (circle, triangle, square, star), and two sizes (regular and small\footnote{A small object has half the height and width of a regular one}).
Following the results of CIVET \cite{rizzoli-etal-2025-civet}, we generate images of $672 \times 672$ pixels, a multiple of the input size of the vision encoder of CLIP, shared across several VLMs.
To capture fine-grained spatial variation, each image is divided into $9\times9$ cells.
Objects are placed in these cells, allowing for more fine-grained analyses on how models' answers about the absolute position vary within the $3\times3$ ground truth regions.
For each combination of attributes, we generate a corresponding VQA sample following the formulation in \Cref{sec:methods:task}.
This process yields 3,888\footnote{Computed as $6 \text{ colors} \times 4 \text{ shapes} \times 2 \text{ sizes} \times 81 \text{ positions (i.e., } 9 \times 9 \text{ cells)}$} balanced image-question pairs.

\textbf{Synthetic Training Set}
We construct a training dataset with the same structure as the evaluation data but distinct color-shape combinations to avoid overlap.
We include the four shapes (\emph{circle}, \emph{triangle}, \emph{square}, \emph{star}) in white and introduce \emph{plus} as an unseen shape in the aforementioned six colors.
This preserves balance across visual attributes while ensuring no color-shape combination is shared between training and testing.
Images follow the same $672\times672$ layout and VQA formulation described in \Cref{sec:methods:task}.
The resulting dataset comprises 1,620\footnote{Computed as $(6 \text{ colored plusses} + 4 \text{ white shapes}) \times 2 \text{ sizes} \times 81 \text{ positions}$} image-question pairs, balanced across all positions.

\textbf{Real-World Dataset}
We construct training and test datasets starting from the train and validation splits of COCO, as test annotation is not provided.
For each image, we generate one or more VQA samples querying the position of a specific object category, e.g., \textit{``Where is the person?''}.
To ensure unambiguous questions, we include only objects that appear as the only instance of their category.
The position of each target object is computed as the center of its bounding box and assigned to one of the nine grid regions defined in \Cref{sec:methods:task}.
We obtain a training set of 201,358 questions and 95,899 images, and an evaluation set of 8,548 questions and 4,109 images.
We split the training set, keeping 80\% (161,086) for training and 20\% (40,272) for validation.

\textbf{Synthetic Set with Distractors}
\label{app:synth_with_distractors_construction}
Real-world scenes often contain multiple objects, many of which are irrelevant to the query. To approximate this complexity and study robustness, we extend the synthetic datasets by adding distractor objects. Each image includes one target object (referenced in the question), and one or more distractors that differ in color, shape, or both. This design allows us to test whether exposure to cluttered visual contexts during fine-tuning improves the model’s ability to attend to task-relevant information. We generate variants containing one, three, or five distractors per image. For images with white target shapes, distractors vary only in shape while retaining the white color; for colored plusses, distractors vary in color while maintaining the same shape. All distractors are placed in random non-overlapping positions within the $9\times9$ cell grid.

\section{Fine-tuning Details}
\label{app:fine_tuning_details}
Each model is fine-tuned using LoRA \cite{hu2022lora} with a rank of 32 and $\alpha$ of 64. Following standard practice and recent findings emphasizing the role of attention in spatial reasoning \cite{chen2025why_spatial_reasoning_hard}, LoRA adapters are applied to the query, key, and value matrices of the attention layers. Fine-tuning is performed for up to 10 epochs with early stopping patience of 2 epochs, a learning rate of $10^{-4}$, using 80\% of the training split for optimization and reserving the remaining 20\% for validation. Models are fine-tuned and tested on the \emph{Absolute Position} task (\cref{sec:methods:task}). Each model is prompted with the image and a closed-ended question, and answers are obtained through greedy decoding. A response is marked as correct only if it contains exactly one of the predefined positional labels.
To reduce output verbosity, which can hinder automatic evaluation, as observed in \cite{rizzoli-etal-2025-civet}, each question is prefixed with the instruction \textit{``Answer with as few words as possible."}, which has been shown to reliably constrain the model’s output to one of the valid options (see attached data samples for examples of the complete prompt).
For CLIP, which follows a dual-encoder architecture, we reformulate the task as an image-text retrieval problem. For each of the nine possible answers, we generate a textual candidate consisting of the same question followed by the position label, encode both the image and the text, and select the answer corresponding to the text representation with the highest cosine similarity to the image embedding.
We fine-tune CLIP starting from the cross-entropy loss used in its original training \cite{radford2021clip}, but only using the component that optimizes for the selection of the correct text candidate given an image.

All experiments were run on a single Nvidia A100 GPU of 80GB.
Following, we report the HuggingFace checkpoints used for each model:
\begin{itemize}
    \item \normalsize \llavas 7B: \small\url{https://huggingface.co/llava-hf/llava-v1.6-vicuna-7b-hf}
    \item \normalsize \llavaone-8B-Instruct: \small\href{https://huggingface.co/lmms-lab/LLaVA-OneVision-1.5-8B-Instruct}{\texttt{https://huggingface.co/lmms-lab/LLaVA-}} \\ \small\href{https://huggingface.co/lmms-lab/LLaVA-OneVision-1.5-8B-Instruct}{\texttt{OneVision-1.5-8B-Instruct}}
    \item \normalsize\molmos 7B-O: \small\url{https://huggingface.co/allenai/Molmo-7B-O-0924}
    \item \normalsize\qwen-7B-Instruct: \small\url{https://huggingface.co/Qwen/Qwen2-VL-7B-Instruct}
    \item \normalsize \clips ViT-L/14-336px: \small\url{https://huggingface.co/openai/clip-vit-large-patch14}
\end{itemize}

\begin{table}[t]
  \centering
  \small
  \setlength{\tabcolsep}{10pt}
  \begin{tabular}{@{}lr@{}}
    \toprule
    Model & \makecell{Accuracy (\%)}  \\
    \midrule
    \multirow{1}{*}{\llava} & 100 {\scriptsize$\pm$1} \textcolor{green!60!black}{($\uparrow$~58)}  \\

    \multirow{1}{*}{\llavaone} & 100 {\scriptsize$\pm$0} \textcolor{green!60!black}{($\uparrow$~33)}  \\

    \multirow{1}{*}{\molmo}  & 96 {\scriptsize$\pm$0} \textcolor{green!60!black}{($\uparrow$~34)}  \\

    \multirow{1}{*}{\qwen}  & 99 {\scriptsize$\pm$0} \textcolor{green!60!black}{($\uparrow$~38)}  \\

    \multirow{1}{*}{\clip}  & 100 {\scriptsize$\pm$0} \textcolor{green!60!black}{($\uparrow$~88)}  \\
    \bottomrule
  \end{tabular}
  \caption{\textbf{Effect of fine-tuning on synthetic data.} Accuracy on the \emph{Absolute Position} task for models fine-tuned and evaluated on the Synthetic Test Set. \textcolor{green!60!black}{($\uparrow$~Value)} shows the absolute improvement with respect to the \textit{Base Model}.
  Fine-tuning leads to near-perfect performance across all models.
  }
  \label{tab:ft_synth}
\end{table}
\begin{table}
    \centering
        \centering
        \begin{adjustbox}{width=\columnwidth}
        \small
        \setlength{\tabcolsep}{5pt}
        \begin{tabular}{@{}llll@{}}
            \toprule
            \multirow{3}{*}{\textbf{Model}} & \multirow{3}{*}{{\makecell[c]{\textbf{Training Set}\\(\#Samples)}}} & \multicolumn{2}{c}{\multirow{2}{*}{\makecell[c]{\textbf{Test Set}\\ \textbf{Accuracy (\%)}}}} \\ \\
            \cmidrule{3-4}
             & & \textbf{Synthetic} & \textbf{COCO} \\
            \midrule

            \multirow{3}{*}{\llava}
              & \textit{Base Model}                     & 42 & 30 \\
              & Synthetic (1.3k)                        & 100 {\scriptsize$\pm$1} & 43 {\scriptsize$\pm$17} \\
              & COCO \textit{Complete} (161k)   & 0 {\scriptsize$\pm$0} & 0 {\scriptsize$\pm$0} \\

            \midrule
            \multirow{3}{*}{\llavaone}
              & \textit{Base Model}                     & 67 & 45 \\
              & Synthetic (1.3k)                        & 100 {\scriptsize$\pm$0} & 65 {\scriptsize$\pm$3} \\
              & COCO \textit{Complete} (161k) & 11 {\scriptsize$\pm$0} & 26 {\scriptsize$\pm$1} \\

            \midrule
            \multirow{3}{*}{\molmo}
              & \textit{Base Model}                     & 62 & 37 \\
              & Synthetic (1.3k)                        & 96 {\scriptsize$\pm$3} & 58 {\scriptsize$\pm$5} \\
              & COCO \textit{Complete} (161k)   & 4 {\scriptsize$\pm$5} & 6 {\scriptsize$\pm$4} \\

            \midrule
            \multirow{3}{*}{\qwen}
              & \textit{Base Model}                     & 61 & 39 \\
              & Synthetic (1.3k)                        & 99 {\scriptsize$\pm$0} & 60 {\scriptsize$\pm$4} \\
              & COCO \textit{Complete} (161k)   & 9 {\scriptsize$\pm$4} & 20 {\scriptsize$\pm$8} \\

            \midrule
            \multirow{3}{*}{\clip}
              & \textit{Base Model}                     & 12 & 22 \\
              & Synthetic (1.3k)                        & 100 {\scriptsize$\pm$0} & 22 {\scriptsize$\pm$9} \\
              & COCO \textit{Complete} (161k)   & 11 {\scriptsize$\pm$0} & 36 {\scriptsize$\pm$0} \\

            \bottomrule
        \end{tabular}
        \end{adjustbox}
    \caption{
        \textbf{Cross-domain transfer to real-world data}.
        Accuracy (\%) on the \emph{Absolute Position} task for models fine-tuned on balanced Synthetic (1.3k), and COCO Complete (161k) test sets.
        $\pm$ denotes the standard deviation obtained from 5 runs.
    }
    \label{tab:synth_vs_coco_train_appendix}
\end{table}

\begin{table}[t]
        \centering
        \small
        \setlength{\tabcolsep}{4pt}
        \begin{tabular}{@{}lp{1.85cm}l@{}}
            \toprule
            \multirow{2}{*}{\textbf{Model}} & \multicolumn{2}{c}{\textbf{Test Set} \textbf{Accuracy (\%)}} \\
            \cmidrule{2-3}
             & \textbf{Synthetic} & \textbf{COCO} \\
            \midrule

            \multirow{1}{*}{\llava}
              & 71 {\scriptsize$\pm$10} & 67 {\scriptsize$\pm$2} \\

            \midrule
            \multirow{1}{*}{\llavaone}
              & 77 {\scriptsize$\pm$6} & 64 {\scriptsize$\pm$3} \\

            \midrule
            \multirow{1}{*}{\molmo}
              & 80 {\scriptsize$\pm$4} & 45 {\scriptsize$\pm$2} \\

            \midrule
            \multirow{1}{*}{\qwen}
              & 80 {\scriptsize$\pm$5} & 61 {\scriptsize$\pm$4} \\

            \midrule
            \multirow{1}{*}{\clip}
              & 13 {\scriptsize$\pm$0} & 36 {\scriptsize$\pm$1} \\

            \bottomrule
        \end{tabular}
        \caption{
            \textbf{Fine-Tuning on balanced real-world data}.
            Accuracy (\%) on the \emph{Absolute Position} task for models fine-tuned on a COCO Subset (1.3k), balanced in terms of object category and position.
            $\pm$ denotes the standard deviation obtained from 5 runs.
        }
        \label{tab:synth_vs_coco_train_appendix_b}
\end{table}

\section{Additional Results}
\label{app:additional_results}

\subsection{Fine-tuning on Synthetic Data}
\label{app:fine_tuning_results}
\Cref{tab:ft_synth} reports the accuracy for models fine-tuned and tested on the synthetic data, averaged over 5 runs.
All models achieve close to perfect accuracy with minimal variations across runs.

\begin{figure*}
    \centering
    \includegraphics[width=1.00\linewidth]{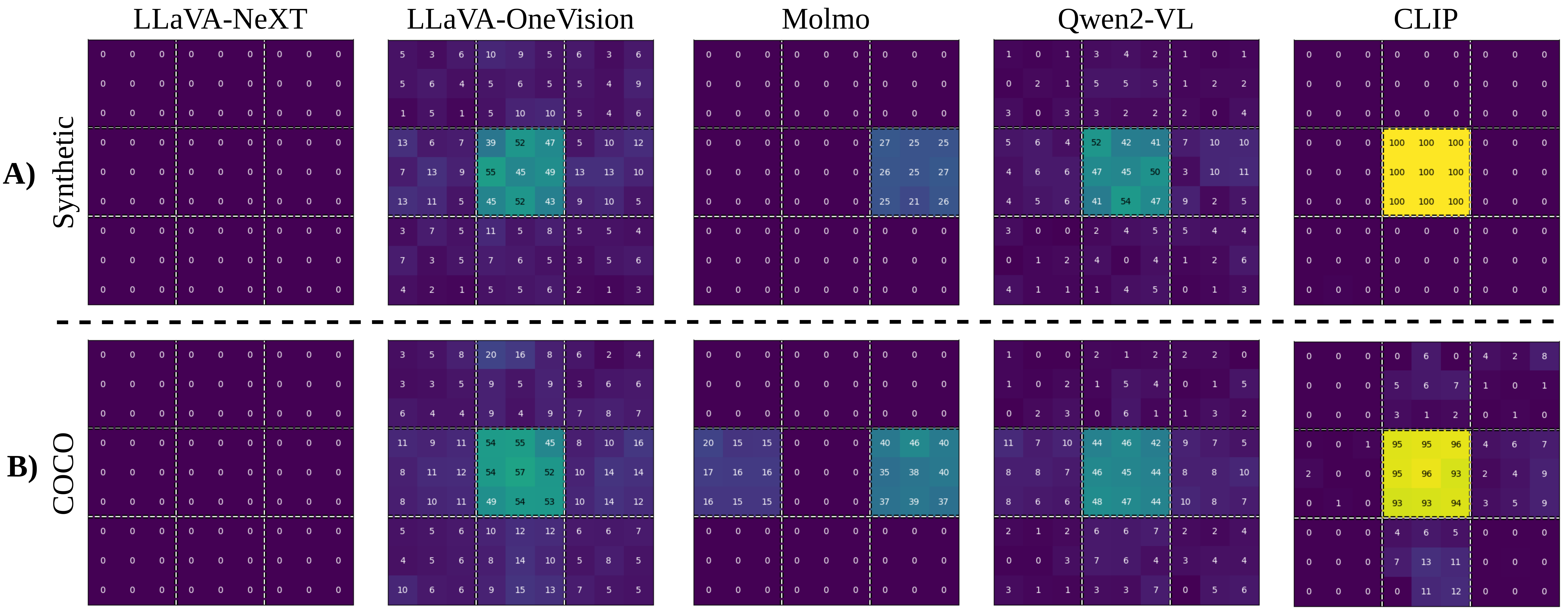}
    \caption{\textbf{Cell-level accuracy of models fine-tuned on the COCO Complete train set} tested on the A) Synthetic and B) COCO test sets.}
    \label{fig:full_coco_appendix}
\end{figure*}

\subsection{Cross-Domain Transfer}
\label{app:synth_to_coco_transfer}
\Cref{tab:synth_vs_coco_train_appendix} reports the accuracy for models fine-tuned on synthetic data and tested on COCO (unmatched setting) and models fine-tuned and tested on COCO (matched setting), extending the results of \Cref{tab:synth_vs_coco_train} (\cref{sec:experiments}) with the standard deviation obtained from five runs.
Similarly, \Cref{tab:synth_vs_coco_train_appendix_b} reports the standard deviation for models fine-tuned on the COCO Subset, balanced in terms of object category and position, extending the results of \Cref{tab:synth_vs_coco_train_b} (\cref{sec:experiments}).

\Cref{fig:full_coco_appendix} reports the cell-level accuracy of models fine-tuned on the complete COCO.
Models show overall low performance in most regions. \clip, \llavaone, and \qwens achieve their best performance in the center, the most common position in the COCO Complete train set.
With the exception of center left and center right for \molmo, models achieve very low performance in all regions outside the center.

\Cref{tab:synth_vs_coco_train_invalid_appendix} reports the percentage of answers that were considered invalid (none of the possible positions, or multiple positions) for the base models and models fine-tuned on the Synthetic and COCO train sets.
When fine-tuned on the Synthetic data or the COCO Subset, models generate invalid answers only rarely.
Fine-tuning on the complete COCO training set degrades the output of the models, with \llavas always generating empty strings, and \molmos often generating tokens from the set \{\textit{center}, \textit{right}, \textit{left}\} up to the token limit, resulting in invalid answers.
\qwens also exhibits some invalid answers due to empty strings or answering with multiple options (e.g., "top left center left").

\begin{table}
    \centering
        \centering
        \begin{adjustbox}{width=\columnwidth}
        \small
        \setlength{\tabcolsep}{5pt}
        \begin{tabular}{@{}llrr@{}}
            \toprule
            \multirow{3}{*}{\textbf{Model}} & \multirow{3}{*}{{\makecell[c]{\textbf{Training Set}\\(\#Samples)}}} & \multicolumn{2}{c}{\multirow{2}{*}{\makecell[c]{\textbf{Test Set}\\ \textbf{Invalid Answers (\%)}}}} \\ \\
            \cmidrule{3-4}
             & & \textbf{Synthetic} & \textbf{COCO} \\
            \midrule

            \multirow{3}{*}{\llava}
              & \textit{Base Model}             & 0.0 & 0.1 \\
              & Synthetic (1.3k)                & 0.0 & 0.0 \\
              & COCO \textit{Complete} (161k) & 100.0 & 100.0 \\

            \midrule

            \multirow{3}{*}{\llavaone}
              & \textit{Base Model}             & 0.0 & 1.3 \\
              & Synthetic (1.3k)                & 0.0 & 0.1 \\
              & COCO \textit{Complete} (161k) & 0.2 & 0.3 \\

            \midrule

            \multirow{3}{*}{\molmo}
              & \textit{Base Model}             & 0.0 & 0.0 \\
              & Synthetic (1.3k)                & 0.0 & 0.1 \\
              & COCO \textit{Complete} (161k) & 52.7 & 45.7 \\

            \midrule

            \multirow{3}{*}{\qwen}
              & \textit{Base Model}             & 0.3 & 1.2 \\
              & Synthetic (1.3k)                & 0.0 & 0.8 \\
              & COCO \textit{Complete} (161k) & 21.0 & 20.4 \\

            \midrule

            \multirow{3}{*}{\clip}
              & \textit{Base Model}             & 0.0 & 0.0 \\
              & Synthetic (1.3k)                & 0.0 & 0.0 \\
              & COCO \textit{Complete} (161k) & 0.0 & 0.0 \\

            \bottomrule
        \end{tabular}
        \end{adjustbox}
    \caption{
        \textbf{Invalid answers in the cross-domain transfer experiments}.
        Percentage (\%) of invalid answers on the \emph{Absolute Position} task, i.e., answers that do not contain exactly one of the nine positions.
    }
    \label{tab:synth_vs_coco_train_invalid_appendix}
\end{table}

\subsection{Cell-level Accuracy}
\label{app:cell_level_accuracy}
\begin{figure}
    \centering
    \includegraphics[width=0.9\linewidth]{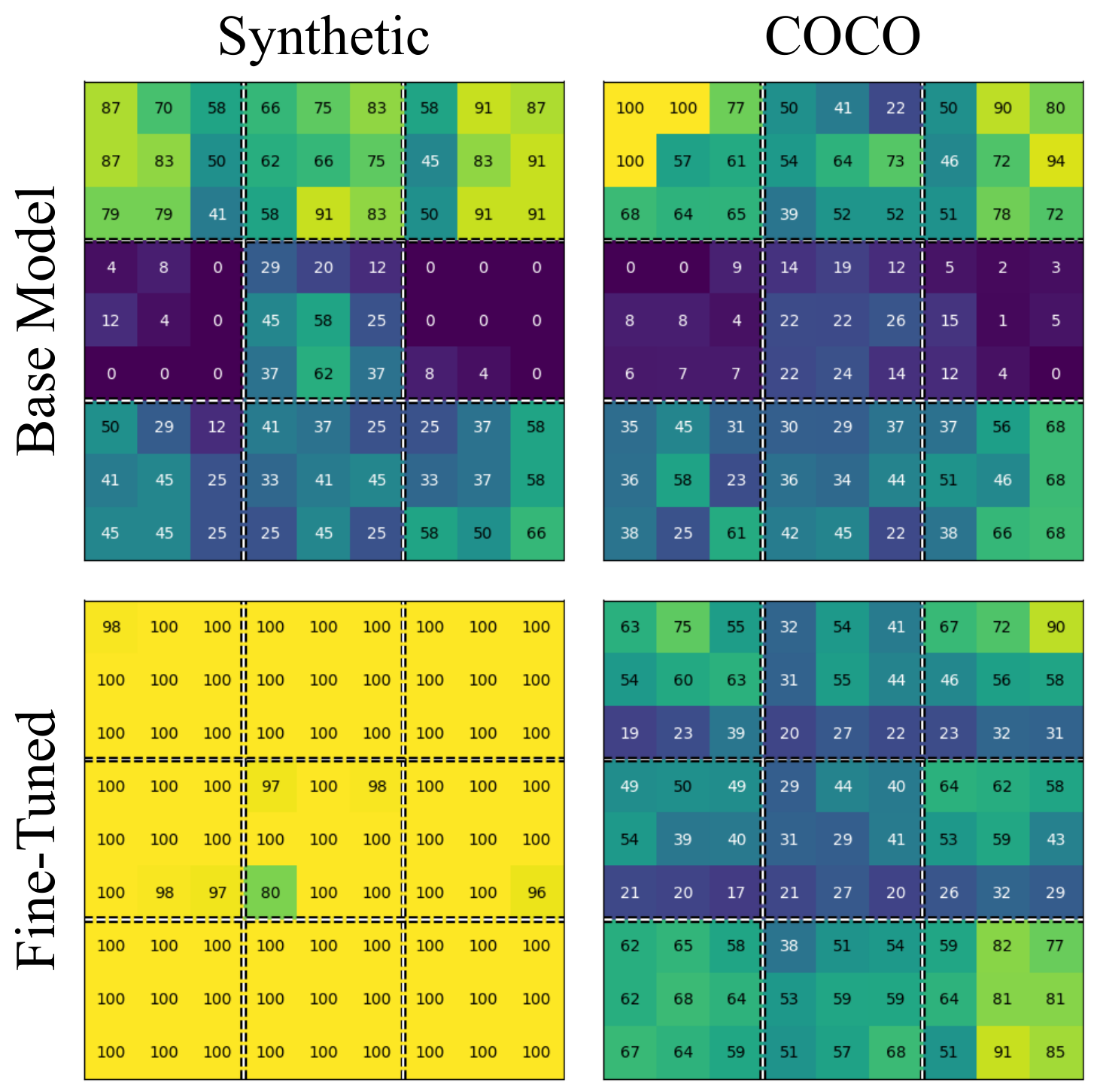}
    \caption{\textbf{Cell-level accuracy of \llavas 7B before and after fine-tuning} on synthetic data, on both the synthetic and COCO Absolute Position test sets.}
    \label{fig:improvements_synth_coco_llava}
\end{figure}

\begin{figure}
    \centering
    \includegraphics[width=0.9\linewidth]{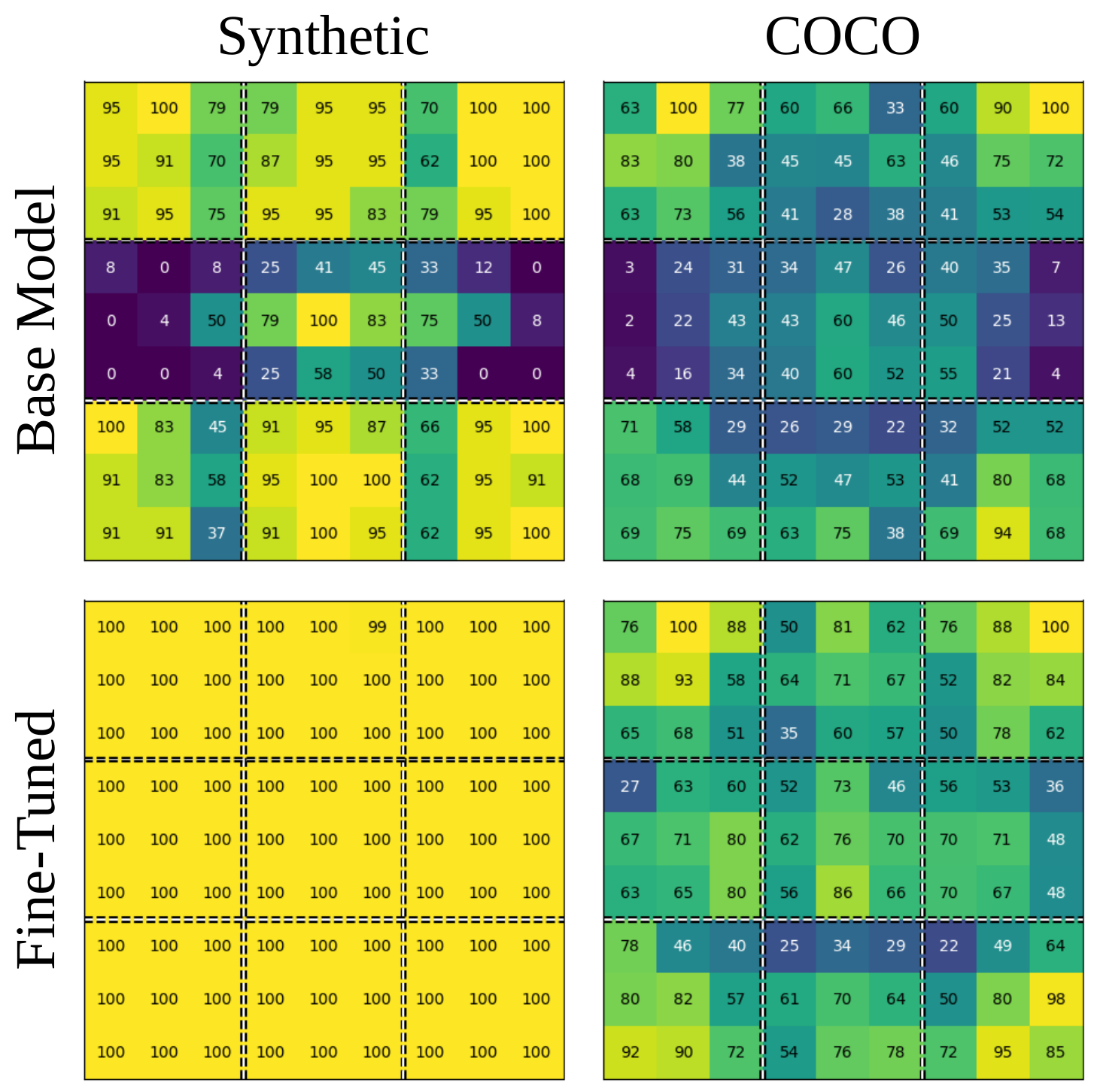}
    \caption{\textbf{Cell-level accuracy of \llavaones 8B before and after fine-tuning} on synthetic data, on both the synthetic and COCO Absolute Position test sets.}
    \label{fig:improvements_synth_coco_llavaones}
\end{figure}

\begin{figure}
    \centering
    \includegraphics[width=0.9\linewidth]{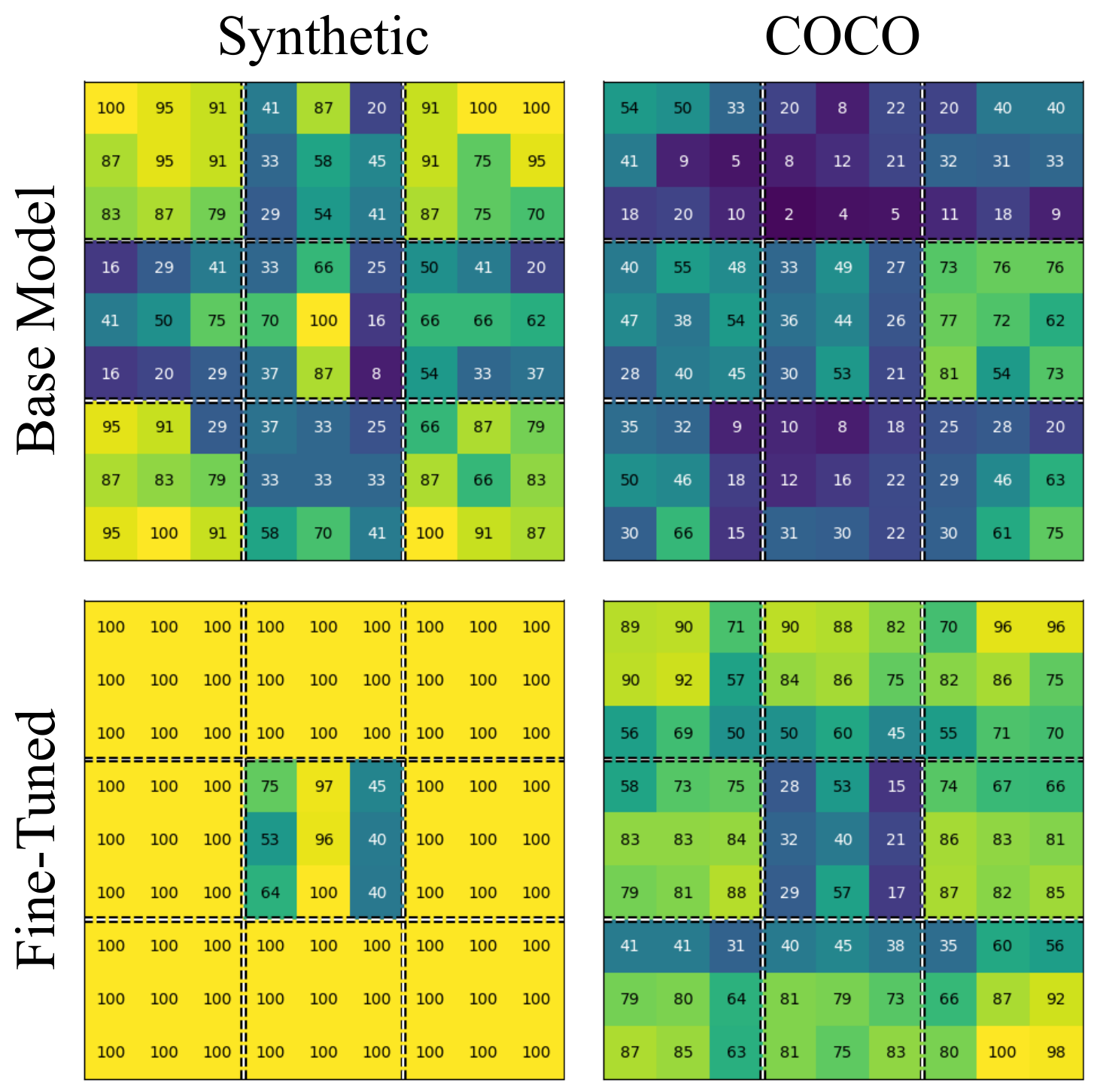}
    \caption{\textbf{Cell-level accuracy of \molmos 7B before and after fine-tuning} on synthetic data, on both the synthetic and COCO Absolute Position test sets.}
    \label{fig:improvements_synth_coco_molmo}
\end{figure}

\begin{figure}
    \centering
    \includegraphics[width=0.9\linewidth]{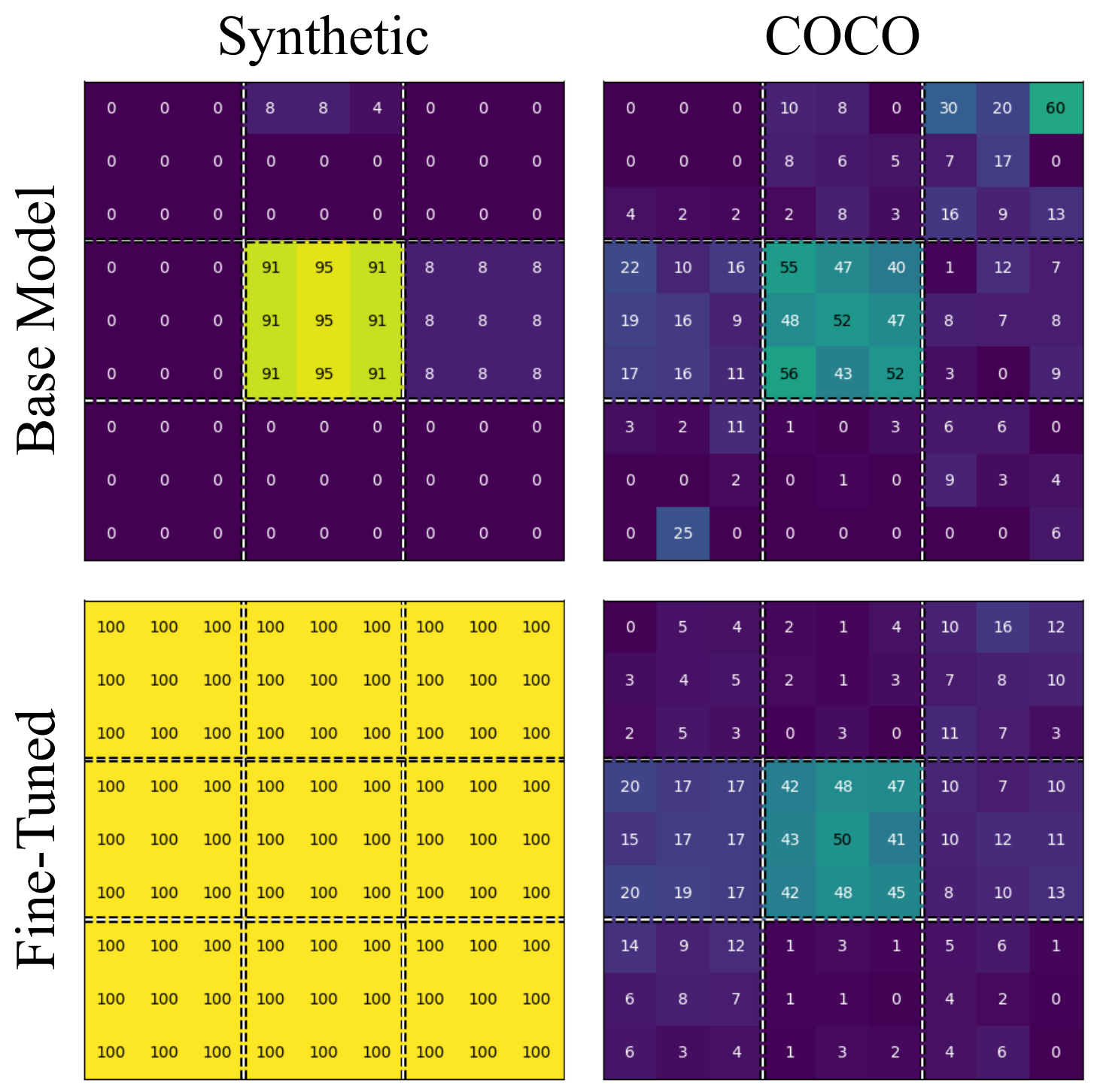}
    \caption{\textbf{Cell-level accuracy of \clips before and after fine-tuning} on synthetic data, on both the synthetic and COCO Absolute Position test sets.}
    \label{fig:improvements_synth_coco_clip}
\end{figure}

We report the cell-level accuracy for \llavas (\cref{fig:improvements_synth_coco_llava}), \llavaones (\cref{fig:improvements_synth_coco_llavaones}), \molmos (\cref{fig:improvements_synth_coco_molmo}), and \clips (\cref{fig:improvements_synth_coco_clip}).
Similarly to \qwens (see \Cref{fig:improvements_synth_coco_qwen} in \cref{sec:experiments}), the encoder-decoder VLMs initially show strong spatial biases and after fine-tuning on synthetic data, the performance on the synthetic set becomes close to uniform.
This is also reflected on the COCO test set, with \llavas and \molmos obtaining most of the improvement where performance was lowest.
Instead, while \clips obtains perfect accuracy across positions on the synthetic test set after fine-tuning, this improvement does not transfer to the COCO test set.

\subsection{Scene Complexity \& Distractors}
\label{app:distractors_results}
In \Cref{tab:distractors_appendix}, we present additional results on increasing scene complexity.
These results extend those in \Cref{tab:distractors} (\cref{sec:ablation_repr_analysis}) by including evaluating on the synthetic test set with the same number of distractors used during fine-tuning and evaluating on the synthetic test with five distractors.
Regardless of fine-tuning data, the encoder-decoder models show a decrease in performance when scene complexity increases.
When tested with five distractors, \molmos and \qwens show little to no benefit from fine-tuning with distractors, while \llavas and \llavaones show a substantial gain with as few as one distractor seen during fine-tuning.
However, adding five distractors to \llavas results in reduced performance on all test sets, suggesting only moderate complexity is beneficial for the model.

\subsection{Layer-Wise Analysis}
\label{app:layerwise_results}
\begin{figure*}
    \centering
    \includegraphics[width=1.00\linewidth]{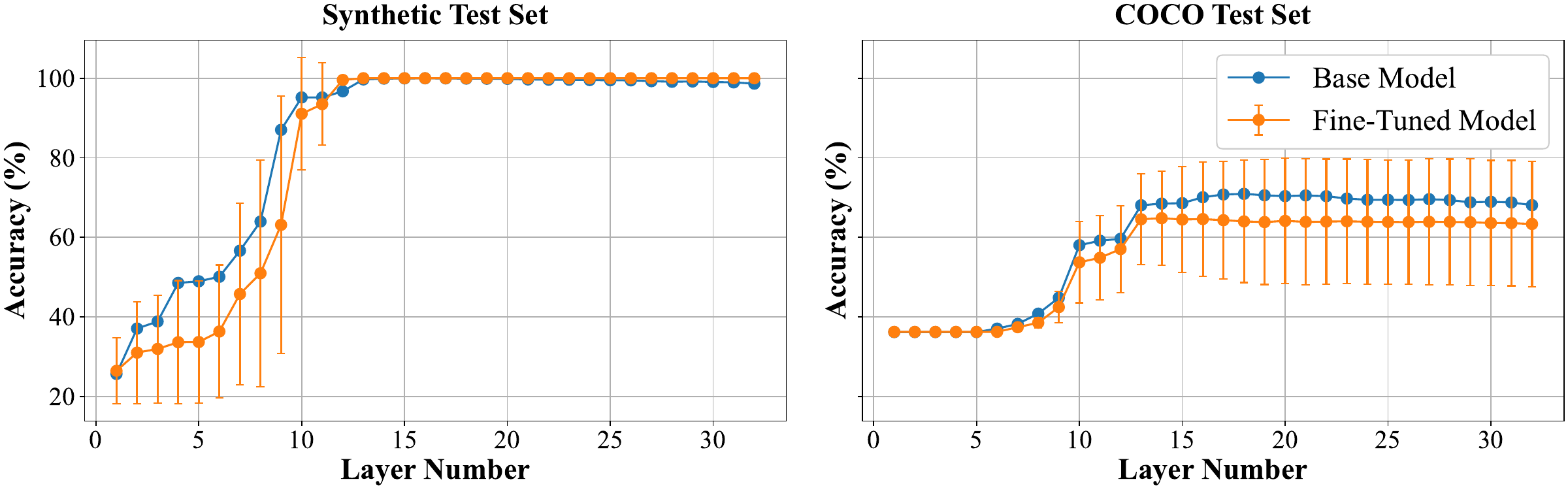}
    \caption{\textbf{Layer-wise probing accuracy of \llavas 7B} before (blue) and after (orange) fine-tuning on the synthetic dataset, evaluated on Synthetic (left) and COCO (right). Error bars represent standard deviation across fine-tuning runs.}
    \label{fig:layerwise_llava}
\end{figure*}

\begin{figure*}
    \centering
    \includegraphics[width=1.00\linewidth]{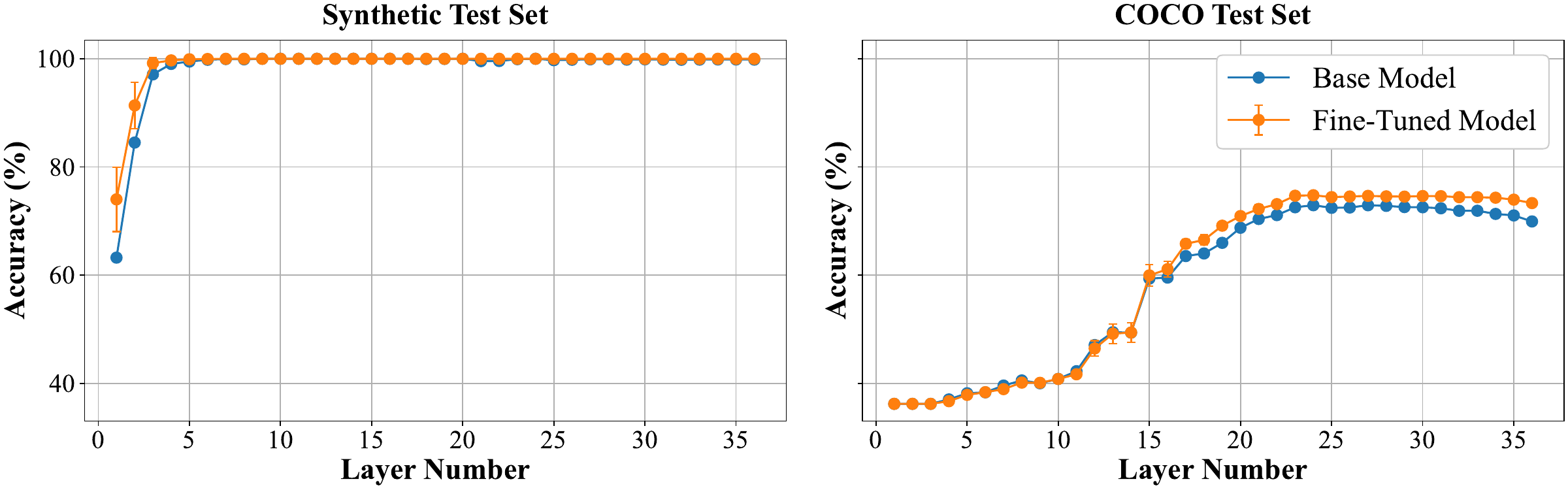}
    \caption{\textbf{Layer-wise probing accuracy of \llavaones 8B} before (blue) and after (orange) fine-tuning on the synthetic dataset, evaluated on Synthetic (left) and COCO (right). Error bars represent standard deviation across fine-tuning runs.}
    \label{fig:layerwise_llavaones}
\end{figure*}

\begin{figure*}
    \centering
    \includegraphics[width=1.00\linewidth]{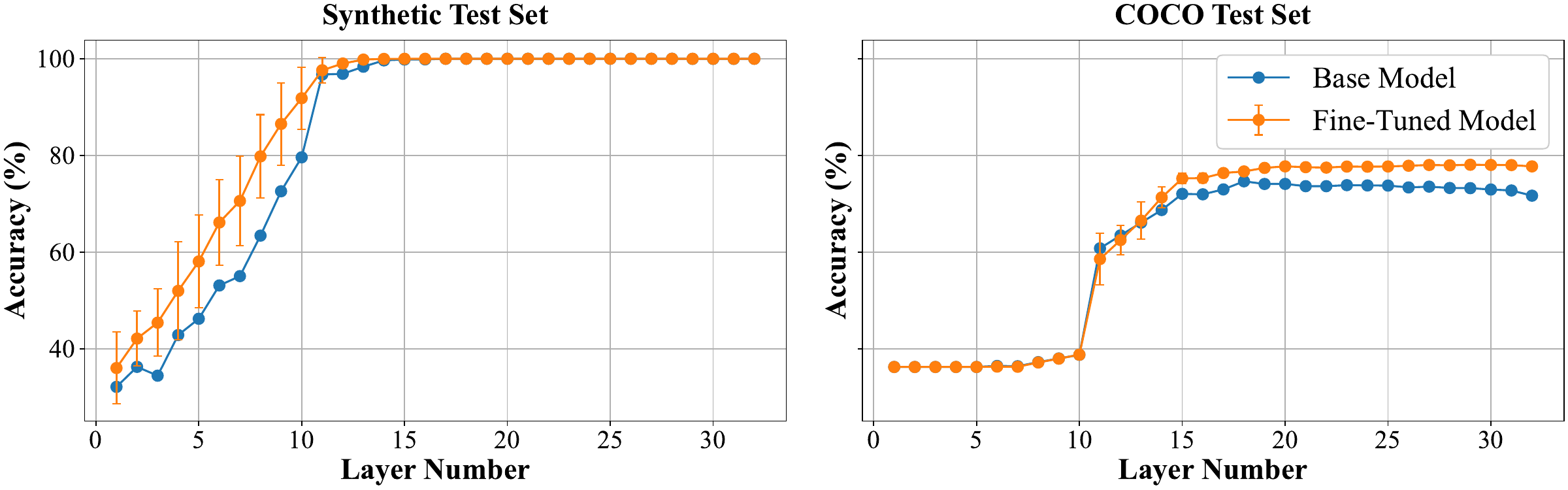}
    \caption{\textbf{Layer-wise probing accuracy of \molmos 7B} before (blue) and after (orange) fine-tuning on the synthetic dataset, evaluated on Synthetic (left) and COCO (right). Error bars represent standard deviation across fine-tuning runs.}
    \label{fig:layerwise_molmo}
\end{figure*}

\begin{figure*}
    \centering
    \includegraphics[width=1\linewidth]{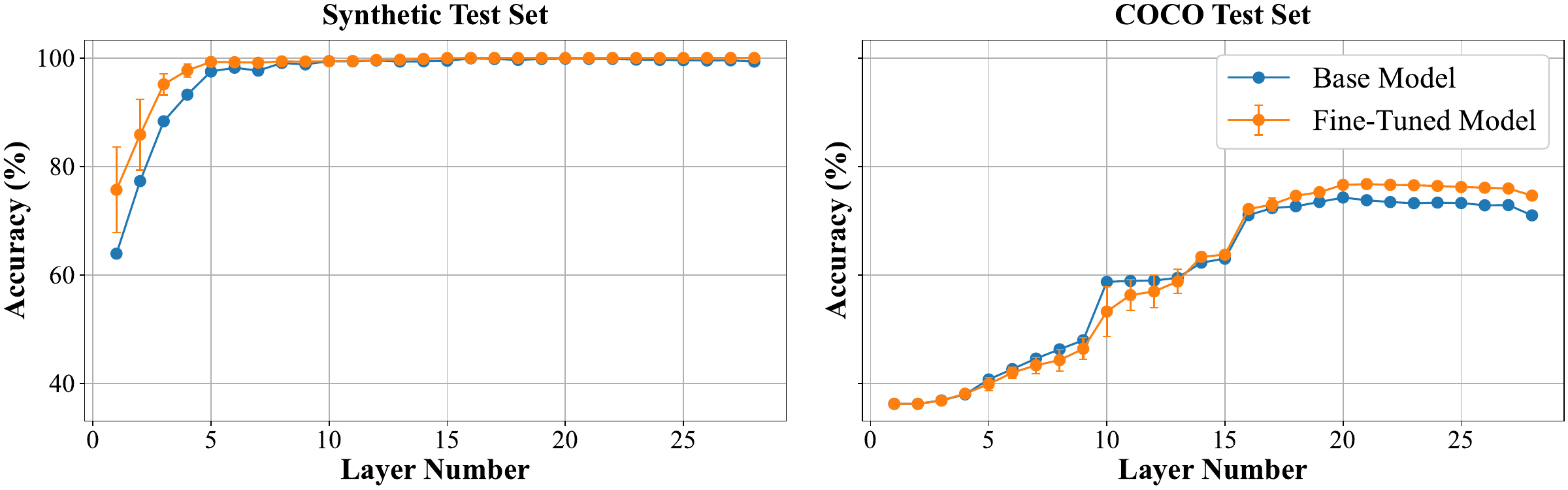}
    \caption{\textbf{Layer-wise probing accuracy of \qwens 7B} before (blue) and after (orange) fine-tuning on the synthetic dataset, evaluated on Synthetic (left) and COCO (right). Error bars represent standard deviation across fine-tuning runs.}
    \label{fig:layerwise_qwen}
\end{figure*}

We report the results for the layer-wise analysis for \llavas (\cref{fig:layerwise_llava}), \llavaones (\cref{fig:layerwise_llavaones}), \molmos (\cref{fig:layerwise_molmo}), and \qwens (\cref{fig:layerwise_qwen}).
The models show a similar trend, rapidly improving performance in the initial layers on synthetic data, while having a slower rise on the more complex scenes of the COCO test set.
For all models, fine-tuning improves the representations for synthetic data.
For \llavaone, \molmo, and \qwen, this improvement transfers to real-world data.
However, \llavas shows mildly reduced performance after fine-tuning, with high variability across runs.
Together with the improvement shown on the Synthetic test, this suggests \llavas is more prone to overfitting on the synthetic data.
This is in line with the experiments on training set scale (\cref{fig:scaling_data_COCO_test} in \cref{sec:experiments}), where \llavas obtains the most transfer after fine-tuning on 10\% of the Synthetic training set, with performance decreasing with larger subsets.

\begin{table*}[t]
  \centering
  \small
      \begin{tabular}{@{}llrrrr@{}}
        \toprule
        \multirow{2}{*}{\textbf{Model}} &  \multirow{2}{*}{\textbf{Training Set}} & \multicolumn{4}{c}{\textbf{Test Set}} \\\cmidrule{3-6}

        & & \textbf{Synthetic} & \textbf{Synth. w. $N$ Distr.} & \textbf{Synth. w. 5 Distr.} & \textbf{COCO} \\
        \midrule
        \multirow{5}{*}{\llava} & \textit{Base Model} & 42 & — & 36 & 30 \\
        & Synthetic (1.3k)             & 100 {\scriptsize$\pm$1} & — & 58 {\scriptsize$\pm$21} & 42 {\scriptsize$\pm$16} \\
        & \quad\textit{with $N=1$ Distr.} & 100 {\scriptsize$\pm$0} & 94 {\scriptsize$\pm$3} & 77 {\scriptsize$\pm$6} & 42 {\scriptsize$\pm$16} \\
        & \quad\textit{with $N=3$ Distr.} & 100 {\scriptsize$\pm$0} & 89 {\scriptsize$\pm$7} & 82 {\scriptsize$\pm$0} & 54 {\scriptsize$\pm$6} \\
        & \quad\textit{with $N=5$ Distr.} & 81 {\scriptsize$\pm$23} & 65 {\scriptsize$\pm$22} & 65 {\scriptsize$\pm$22} & 48 {\scriptsize$\pm$21} \\
        \midrule

        \multirow{5}{*}{\llavaone} & \textit{Base Model} & 67 & — & 64 & 45 \\
        & Synthetic (1.3k)                & 100 {\scriptsize$\pm$ 0} & — & 89 {\scriptsize$\pm$ 3} & 65 {\scriptsize$\pm$ 3} \\
        & \quad\textit{with $N=1$ Distr.} & 100 {\scriptsize$\pm$ 0} & 100 {\scriptsize$\pm$ 1} & 98 {\scriptsize$\pm$ 2} & 66 {\scriptsize$\pm$ 3} \\
        & \quad\textit{with $N=3$ Distr.} & 100 {\scriptsize$\pm$ 0} & 99 {\scriptsize$\pm$ 1} & 98 {\scriptsize$\pm$ 1} & 60 {\scriptsize$\pm$ 6} \\
        & \quad\textit{with $N=5$ Distr.} & 100 {\scriptsize$\pm$ 0} & 98 {\scriptsize$\pm$ 1} & 98 {\scriptsize$\pm$ 1} & 60 {\scriptsize$\pm$ 9} \\
        \midrule

        \multirow{5}{*}{\molmo} & \textit{Base Model} & 62 & — & 59 & 39 \\
        & Synthetic (1.3k)             & 96 {\scriptsize$\pm$3} & — & 93 {\scriptsize$\pm$2} & 57 {\scriptsize$\pm$5} \\
        & \quad\textit{with $N=1$ Distr.} & 96 {\scriptsize$\pm$2} & 95 {\scriptsize$\pm$2} & 91 {\scriptsize$\pm$3} & 60 {\scriptsize$\pm$2} \\
        & \quad\textit{with $N=3$ Distr.} & 95 {\scriptsize$\pm$3} & 93 {\scriptsize$\pm$4} & 92 {\scriptsize$\pm$5} & 60 {\scriptsize$\pm$6} \\
        & \quad\textit{with $N=5$ Distr.} & 97 {\scriptsize$\pm$2} & 92 {\scriptsize$\pm$2} & 92 {\scriptsize$\pm$2} & 65 {\scriptsize$\pm$1} \\
        \midrule
        \multirow{5}{*}{\qwen} & \textit{Base Model} & 61 & — & 53 & 38 \\
        & Synthetic (1.3k)             &  99 {\scriptsize$\pm$0} & — & 92 {\scriptsize$\pm$8} & 58 {\scriptsize$\pm$4} \\
        & \quad\textit{with $N=1$ Distr.} & 98 {\scriptsize$\pm$2} & 98 {\scriptsize$\pm$2} & 93 {\scriptsize$\pm$4} & 59 {\scriptsize$\pm$3} \\
        & \quad\textit{with $N=3$ Distr.} & 93 {\scriptsize$\pm$5} & 93 {\scriptsize$\pm$4} & 92 {\scriptsize$\pm$5} & 58 {\scriptsize$\pm$4} \\
        & \quad\textit{with $N=5$ Distr.} & 90 {\scriptsize$\pm$3} & 88 {\scriptsize$\pm$7} & 88 {\scriptsize$\pm$7} & 54 {\scriptsize$\pm$3} \\
        \midrule
        \midrule
        \multirow{5}{*}{\clip} & \textit{Base Model} & 12 & — & 11 & 22 \\
        & Synthetic (1.3k)             & 100 {\scriptsize$\pm$0} & — & 15 {\scriptsize$\pm$0} & 22 {\scriptsize$\pm$9} \\
        & \quad\textit{with $N=1$ Distr.} & 25 {\scriptsize$\pm$30} & 25 {\scriptsize$\pm$30} & 16 {\scriptsize$\pm$10} & 14 {\scriptsize$\pm$5} \\
        & \quad\textit{with $N=3$ Distr.} & 11 {\scriptsize$\pm$0} & 11 {\scriptsize$\pm$0} & 11 {\scriptsize$\pm$0} & 22 {\scriptsize$\pm$10} \\
        & \quad\textit{with $N=5$ Distr.} & 11 {\scriptsize$\pm$0} & 11 {\scriptsize$\pm$0} & 11 {\scriptsize$\pm$0} & 28 {\scriptsize$\pm$2} \\
        \bottomrule
      \end{tabular}
  \caption{\textbf{Effect of Distractors on the \emph{Absolute Position} task} for VLMs fine-tuned on the synthetic dataset, when evaluated on Synthetic (no distractors), Synthetic with the same number of Distractors as fine-tuning, Synthetic with 5 Distractors, and on COCO. Results show the average accuracy (\%) across five runs, and $\pm$ denotes the standard deviation.}
  \label{tab:distractors_appendix}
\end{table*}

\end{document}